\documentclass{article} %
\usepackage{iclr2027_conference,times}
\usepackage[T1]{fontenc}

\usepackage{amsmath,amsfonts,bm}

\def\eqref#1{equation~\ref{#1}}
\def\1{\bm{1}}

\DeclareMathAlphabet{\mathsfit}{\encodingdefault}{\sfdefault}{m}{sl}
\SetMathAlphabet{\mathsfit}{bold}{\encodingdefault}{\sfdefault}{bx}{n}

\usepackage{hyperref}
\usepackage{url}
\usepackage{graphicx}
\usepackage{booktabs}
\usepackage{xcolor}
\usepackage{float}
\usepackage{placeins}

\newcommand{\slot}[2]{\if\relax\detokenize{#2}\relax\errmessage{Empty number slot #1}\else#2\fi}
\newcommand{\fslot}[2]{\if\relax\detokenize{#2}\relax\errmessage{Empty number slot #1}\else#2\fi}

\title{A Dominant Supplier Slows Recursive Drift\\More Than It Steers It}

\author{Yangze Liu \\
Shandong University \\
\texttt{yangze2@illinois.edu}
\And
Zhongyi Han\thanks{Corresponding author.} \\
Shandong University \\
\texttt{zhongyi.han@sdu.edu.cn}}

\iclrfinalcopy

\begin{document}

\maketitle

\begin{abstract}
More and more of the text future language models learn from is written by a few of today's models. If one supplier writes most of a shared corpus, does it pull the models trained on it toward its own writing, or change how fast they drift? We retrain eight open models from their base weights on a shared pool of each other's text for five generations, varying the part written by one model, Phi-2, from an equal share to 90\%. The models drift together toward a style with fewer function words, and none starts repeating itself. No share of Phi-2 brings the other models closer to its text than the equal share does. We split each ecosystem's separation from the equal-share one into a delay along its route and a departure from that route, both counted beyond the difference between two equal-share runs. With Phi-2 at 90\%, delay outweighs departure 72 to 28 and 64 to 36 in two runs, and the ecosystem falls \slot{path.lag90}{2.7} and 2.5 generations behind. With Phi-2 at half the pool the two parts are about equal. When SmolLM2 or Qwen3-1.7B writes half instead, the ecosystem slows less or not at all. The departure leans toward Phi-2 more as its share grows, but more than toward every other model only at 90\%. Human text filling a quarter or half of the pool slows the models along the same route.
\end{abstract}

\section{Introduction}
\label{sec:intro}

Language models now write a large part of what is published online. When an industry crawler examined 900{,}000 English web pages that it first found in April 2025, its detector flagged 74.2\% of them as containing some AI-generated text \citep{law2025ai}. However carefully the next generation of models filters its training data, some of this text will be learned from, and the models that learn from it will in turn write text for the generation after them. Studies of this loop have mostly followed a single model trained repeatedly on its own samples, whose output narrows as the tails of its distribution thin out, a process known as model collapse \citep{shumailov2024ai,alemohammad2024selfconsuming}.

The real loop does not run through one model. Text on the web comes from many models, and they are not equally present. An industry survey of 495 US enterprises estimates that three providers take 88\% of enterprise spending on language-model APIs and that one provider holds 54\% of the coding market \citep{menlo2025state}. Economists and competition authorities describe the same concentration in the market for foundation models \citep{korinek2025concentrating,cma2024ai}. If the text that trains the next models follows the market, most of it will come from a handful of suppliers.

Concentration raises a worry that studies of a single model cannot address. If one supplier writes most of the shared corpus, every later model learns mostly from that supplier, and the ecosystem as a whole may be pulled toward the way that supplier writes. A second version of the worry concerns speed. A corpus drawn from fewer sources is less varied, and recursive training on it might wear the ecosystem down faster. Both versions treat a supplier's share of the corpus as a lever, one on where the ecosystem goes and one on how fast it gets there.

Studies of recursive training among several models have so far given every model the same share. Models that exchange text through a shared memory converge in their outputs \citep{wang2025llm}, and so do models trained recursively on each other's outputs \citep{vu2025what}. \citet{vu2025what} raise the case of one dominant provider and set it aside as reducing to a single model that trains on itself (their Appendix~B). \citet{hodel2025epistemic} vary how a fixed corpus is divided among models that each train on their own part, so that concentration there means fewer models with larger parts. The range in which one supplier writes a quarter, a half or nine tenths of a corpus that every model trains on, while the others still contribute, is the range these market shares point to, and it has not been measured before.

We measure it in a controlled ecosystem. Eight open base models from six organizations form a shared ecosystem. In each generation every model writes text, the texts are mixed into a pool of fixed size according to preset shares, and every model is fine-tuned on that pool from its base weights. We compare a uniform ecosystem, in which every model writes an equal share, with ecosystems in which one model, Phi-2, writes 28\%, 50\% or 90\% of the pool, two ecosystems in which SmolLM2 or Qwen3-1.7B writes half instead, and two in which human text fills a quarter or half of the pool. A frozen text encoder tracks each member's own output, and word counts of the same output show what the encoder's distances correspond to in the text.

The drift in these ecosystems is a change of style. In every configuration the members write fewer function words and longer words in each generation, and no member starts repeating itself. No share of Phi-2 brings the other members closer to its text than the equal share does. What a dominant supplier can do to such a drift is bend the shared route toward its own text or change the pace along it, and we measure both on one scale by splitting the separation between two ecosystems into a delay along a common route and a departure from it. Counted beyond the difference between two runs of the uniform ecosystem, most of what Phi-2 adds at nine tenths of the pool is delay in both of our runs, and at half the pool delay and departure are of similar size. The departures lean toward Phi-2's text more than toward any other member's only at nine tenths. Where a dominant supplier changes the drift at all, what it changes most clearly is the pace.

\section{Related Work}
\label{sec:related}

Recursive training on generated data was first studied with a single model that learns from its own output, where the tails of the distribution disappear first and quality declines after them \citep{shumailov2024ai,alemohammad2024selfconsuming}. Later work found conditions that soften the decline. Accumulating data across generations instead of replacing it avoids collapse \citep{gerstgrasser2024model}, and training each generation on a fixed-size sample of the accumulated data produces a slow and gradual degradation \citep{kazdan2024collapse}. Iterative retraining stays stable when enough real data remains in the mix \citep{bertrand2024stability}, and mixing human and synthetic data changes the scaling laws that a model follows \citep{dohmatob2024tale}. Other studies trace the loss of lexical diversity over generations \citep{guo2024curious,briesch2023large}, estimate how much synthetic data a model can absorb before it collapses \citep{seddik2024bad}, and counter the decline with verification or self-correction \citep{feng2024beyond,gillman2024selfcorrecting}. \citet{wu2024generative} find that a model's output is already less diverse than its training data, which they call generative monoculture, and \citet{schaeffer2025position} argue that the term model collapse covers several distinct definitions. In all of these one model learns from its own output.

Several studies let models feed each other. Models that share a retrieval memory converge in their outputs \citep{wang2025llm}, and so do models trained recursively on each other's outputs \citep{vu2025what}, in both cases with every model contributing equally. \citet{hodel2025epistemic} split a fixed corpus among models that each train on their own part and find that concentrating it in fewer models accelerates collapse. \citet{wu2026when} model who trains on whose output as a directed graph and prove that the structure of that graph determines which models collapse (their Section~2). These studies establish that a population of models converges or collapses. We hold the population fixed and ask whether the share of the pool that one supplier writes changes the direction or the speed of its drift.

Concentration in foundation-model markets is studied in economics and competition policy, where homogenization concerns prices and market entry \citep{korinek2025concentrating,cma2024ai,turegeldinova2025ai}. Work on algorithmic monoculture studies homogenization at deployment, when many decision makers rely on the same model \citep{kleinberg2021algorithmic,bommasani2022picking}. Neither line concerns how models change when they train on each other's text.

\begin{figure}[t]
\centering
\includegraphics[width=\textwidth,trim=0 22 0 25,clip]{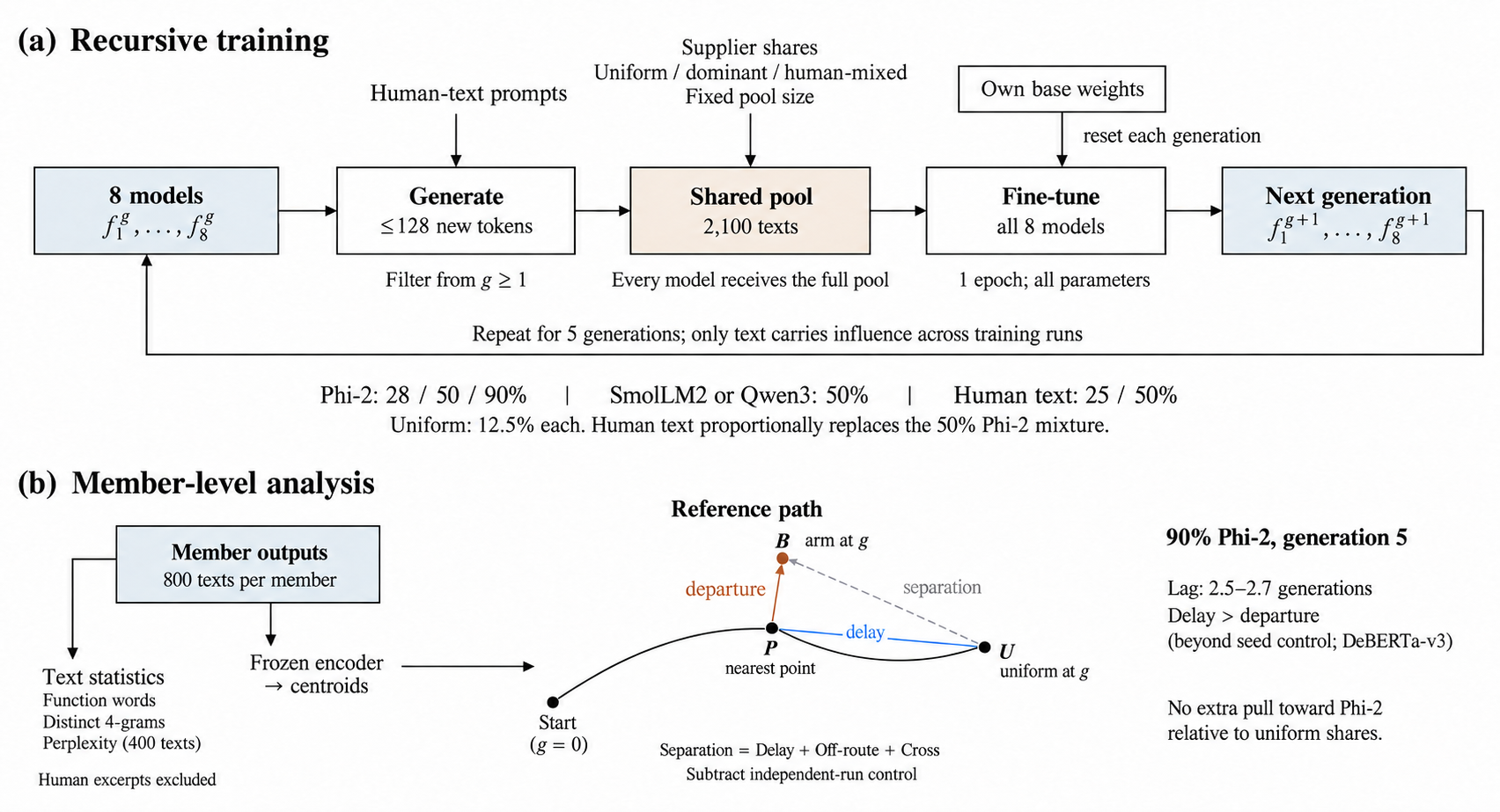}
\caption{Overview. (a)~One generation of the recursive training loop. (b)~The measurements on the members' texts, with a schematic, not drawn from data, of how a separation splits along the reference path.}
\label{fig:pipeline}
\end{figure}

\section{Setup}
\label{sec:setup}

\paragraph{The loop.} An ecosystem has $K=8$ member models (Figure~\ref{fig:pipeline}). In each generation every member writes continuations of short prompts cut from human text, the continuations are mixed into one pool of 2{,}100 texts according to preset shares, and every member is fine-tuned on the whole pool, with all parameters updated, starting again from its own base weights. The fine-tuned members write the next generation's text. Only text crosses generations, and the pool has the same size in every configuration, so configurations differ only in who wrote the pool. We run five generations and call one configuration, together with the ecosystem it produces, an arm.

\paragraph{Members and arms.} The members are eight open base models with 1B to 3B parameters from six organizations: Qwen3-1.7B, Qwen2.5-1.5B, SmolLM2-1.7B, SmolLM3-3B, OLMo-2-1B, StableLM-2-1.6B, Phi-2 and Falcon3-1B (Table~\ref{tab:models} gives the full identifiers). We call the member with the largest share the head. We chose Phi-2 as the head because, under our encoder, its generation-0 text lies farthest from the other members' text, so a pull toward it would be the easiest to see. The uniform arm gives every member 12.5\% of the pool. The 28\%, 50\% and 90\% arms give Phi-2 that share and split the rest evenly among the other seven members. The SmolLM2 and Qwen3 arms give 50\% instead to SmolLM2, whose generation-0 text lies close to the other members' text, or to Qwen3-1.7B, whose text lies about four fifths as far from theirs as Phi-2's. The two human arms start from the 50\% arm and fill a quarter or half of the pool with excerpts of human-written web text and books \citep{gao2020pile}, drawn afresh each generation, which shrinks every member's part in proportion, so Phi-2 writes 37.5\% or 25\% of the pool.

\paragraph{Seeds and pairing.} Every generation request carries its own sampling seed, set by the member, the generation, the prompt and a chain seed. Arms that share a chain seed start from the same generation-0 text and later draw the same prompts with the same seeds, so a comparison between them is paired. The uniform, 28\%, 50\%, 90\%, SmolLM2 and Qwen3 arms run with chain seeds 42 and 43, and the human arms with seed 42. The uniform arm with seed 42 also runs on to generation 10. Two runs of one arm with different chain seeds differ in every sampling seed, and their distance is the seed floor against which we read differences between arms.

\paragraph{Distances.} A frozen DeBERTa-v3 encoder \citep{he2023debertav3,laurer2022less} embeds each text. Under it the 50\% arm lies 5.0 and 6.9 times as far from the uniform arm as two runs of the uniform arm lie from each other, in the two chain seeds, against 2.2 and 2.6 times under the sentence encoder all-mpnet-base-v2 (Table~\ref{tab:rev_encoder2}), so it separates arms more sharply from seed noise, and Section~\ref{sec:direction} compares the two.
Member $m$ of arm $A$ at generation $g$ is summarized by the unit-length mean embedding $c^A_{m,g}$ of its first 800 texts, written $c^{A,s}_{m,g}$ when the chain seed $s$ matters, and human texts never enter these centroids. All arms of a chain seed share generation 0, so the starting centroid $c_{m,0}$ carries no arm index. With $d(x,y)=1-x^\top y$ the cosine distance between unit vectors, the drift of arm $A$, the separation between arms $A$ and $B$, the seed floor of arm $A$ run with chain seeds $s$ and $s'$, and the spread of arm $A$ are averages over the members,
\begin{equation}
\begin{gathered}
D^A(g)=\frac{1}{K}\sum_{m} d\big(c^A_{m,g},c_{m,0}\big),\qquad
S^{AB}(g)=\frac{1}{K}\sum_{m} d\big(c^A_{m,g},c^B_{m,g}\big),\\
F^{A}(g)=\frac{1}{K}\sum_{m} d\big(c^{A,s}_{m,g},c^{A,s'}_{m,g}\big),\qquad
W^{A}(g)=\frac{2}{K(K-1)}\sum_{m<m'} d\big(c^A_{m,g},c^A_{m',g}\big).
\end{gathered}
\label{eq:dist}
\end{equation}
Every distance in the text, figures and tables is a cosine distance in units of $10^{-3}$. A pool can also be summarized by the mean embedding of its 2{,}100 texts, which weights each member by its share (Table~\ref{tab:rev_pool}), and we use this pool scale only for how far a share moves the starting pool.

\paragraph{Direction and pace.} Two cosines describe direction, each built from drift vectors $c^A_{m,g}-c_{m,0}$. The direction cosine $\kappa^{AB}_m(g)$ is the cosine between member $m$'s drift vectors in arms $A$ and $B$, and its mean $\kappa^{AB}(g)$ over the members is one when two arms move every member the same way. The co-movement cosine $\gamma^A_m(g)$ is the cosine between member $m$'s drift vector and the mean drift vector of the other seven members of arm $A$, and its mean $\gamma^A(g)$ is one when every member moves with the rest of its ecosystem. To tell a different route from the same route at a different speed, we join member $m$'s centroids $c^A_{m,0},\dots,c^A_{m,5}$ into a piecewise-linear path $p^A_m(t)$, $0\le t\le 5$, renormalized to unit length, so that $p^A_m(g)=c^A_{m,g}$ at whole generations. Each member of arm $B$ is placed at the position $t_m(g)$ of its nearest point on that path, and the effective generation and the off-route part of the separation are
\begin{equation}
\tau^{B|A}(g)=\frac{1}{K}\sum_{m} t_m(g),\qquad
O^{B|A}(g)=\frac{1}{K}\sum_{m} d\big(c^B_{m,g},\,p^A_m(t_m(g))\big).
\label{eq:path}
\end{equation}
Because $d$ is half a squared Euclidean distance, $S^{AB}(g)$ is the sum of three parts. With $p_m=p^A_m(t_m(g))$ the nearest point, the off-route part is $O^{B|A}(g)$, the delay part $\frac{1}{K}\sum_m d\big(p_m,c^A_{m,g}\big)$ is the chord from the nearest point to arm $A$'s own position, and the cross term is the mean over the members of $(c^B_{m,g}-p_m)^\top(p_m-c^A_{m,g})$. The cross term would vanish on a straight route but not on a bending one, and it is negative when members leave the route toward arm $A$'s position, so the delay part can exceed the separation. We call the vector from $p_m$ to $c^B_{m,g}$ the member's departure. A supplier steers when the departures point toward the head, measured by the cosine between each member's departure and the vector from $p_m$ to the head's current centroid, averaged over the seven other members, and it pulls when the other members' mean distance to the head falls more than in the uniform arm. Because the path ends at generation 5, a run that keeps pace with arm $A$ still projects slightly short of the end and slightly off the path, so we read arm $B$ against a control, a second run $A'$ of arm $A$ with the other chain seed placed on the same path. The lag is $\tau^{A'|A}(g)-\tau^{B|A}(g)$, and the parts beyond the control subtract those of $A'$.

\paragraph{Text statistics.} The function-word rate counts occurrences of the 20 most frequent English function words (the, of, and, a, to, in, is, that, it, for, was, with, as, on, be, at, by, this, from, or) per 1{,}000 lowercase words and is averaged over the members. It falls in every generation of the uniform arm, which makes it a clock that does not depend on the encoder. An arm's generation-5 rate, placed by linear interpolation on the uniform arm's curve with the same chain seed, gives the generation at which the uniform arm wrote at that rate. The perplexity of the members' texts under GPT-2-large \citep{radford2019language}, averaged over the members other than the head, rises in every generation and gives a second clock built the same way. The distinct 4-gram fraction of a member is the number of distinct word 4-grams in its texts divided by the total number, and it falls whenever a member repeats itself.

\section{Results}
\label{sec:results}

We first read what the drift is in the members' text, then ask whether a dominant supplier changes its direction or its pace. Table~\ref{tab:split} splits the separation of every arm into these two parts.

\begin{table}[t]
\caption{Separation $S$ from the reference arm at generation 5 and its three parts (Section~\ref{sec:setup}), in units of $10^{-3}$. Each member is placed on its own path in the reference arm with the same chain seed. The control is a second run of the reference arm with the other chain seed (the rows without a lag), $S/F$ divides $S$ by the control's, and the delay share is the delay's percentage of delay plus off-route. $^{\dagger}$No share is given where the off-route part lies below the control's. $^{\ddagger}$Seven and eight of the eight members of the Qwen3 arm lie at the end of the path in its two runs, so its split is not given and its lag understates its lead. Toward gives the departure cosine toward the head (Phi-2, or the member named in the row) and the largest such cosine toward another member (Table~\ref{tab:towardnull}).}
\label{tab:split}
\centering
\footnotesize
\setlength{\tabcolsep}{2.2pt}
\begin{tabular}{lrrrrrrrrrrrr}
\toprule
 & & & \multicolumn{3}{c}{Parts of $S$} & \multicolumn{2}{c}{Beyond control} & \multicolumn{2}{c}{Delay share (\%)} & \multicolumn{2}{c}{Toward} & \\
\cmidrule(lr){4-6}\cmidrule(lr){7-8}\cmidrule(lr){9-10}\cmidrule(lr){11-12}
Arm, seed & $S$ & $S/F$ & Delay & Off & Cross & Delay & Off & Raw & Beyond & Head & Other & Lag \\
\midrule
\multicolumn{13}{l}{\emph{Reference: the uniform arm with the same chain seed}} \\
Uniform, 43 & 0.79 & 1.0 & 0.12 & 0.67 & 0.00 &  &  & 15 &  & 0.03 & 0.61 &  \\
Uniform, 42 & 0.79 & 1.0 & 0.09 & 0.70 & 0.00 &  &  & 12 &  & 0.24 & 0.53 &  \\
28\% Phi-2, 42 & 1.07 & 1.4 & 0.17 & 0.89 & 0.01 & 0.05 & 0.22 & 16 & 20 & 0.36 & 0.66 & 0.11 \\
28\% Phi-2, 43 & 2.32 & 2.9 & 0.70 & 1.51 & 0.11 & 0.61 & 0.82 & 32 & 43 & 0.54 & 0.74 & 0.67 \\
50\% Phi-2, 42 & 3.95 & 5.0 & 1.68 & 1.98 & 0.28 & 1.57 & 1.31 & 46 & 54 & 0.61 & 0.63 & 0.75 \\
50\% Phi-2, 43 & 5.45 & 6.9 & 2.40 & 2.47 & 0.59 & 2.31 & 1.77 & 49 & 57 & 0.59 & 0.58 & 1.28 \\
90\% Phi-2, 42 & 24.61 & 31.0 & 19.22 & 7.96 & $-$2.57 & 19.10 & 7.29 & 71 & 72 & 0.72 & 0.66 & 2.66 \\
90\% Phi-2, 43 & 22.82 & 28.8 & 13.39 & 8.22 & 1.21 & 13.30 & 7.52 & 62 & 64 & 0.73 & 0.70 & 2.48 \\
50\% SmolLM2, 42 & 1.48 & 1.9 & 0.16 & 1.32 & 0.00 & 0.04 & 0.64 & 11 & 6 & 0.38 & 0.66 & 0.07 \\
50\% SmolLM2, 43 & 3.91 & 4.9 & 0.41 & 3.28 & 0.21 & 0.32 & 2.58 & 11 & 11 & 0.48 & 0.79 & 0.50 \\
50\% Qwen3, 42 & 3.54 & 4.5 & \multicolumn{7}{c}{$\ddagger$} & 0.78 & 0.78 & $-$0.23 \\
50\% Qwen3, 43 & 2.72 & 3.4 & \multicolumn{7}{c}{$\ddagger$} & 0.76 & 0.79 & $-$0.19 \\
25\% human, 42 & 6.12 & 7.7 & 5.46 & 0.90 & $-$0.24 & 5.34 & 0.22 & 86 & 96 & 0.42 & 0.46 & 1.54 \\
50\% human, 42 & 13.56 & 17.1 & 14.31 & 0.91 & $-$1.67 & 14.19 & 0.24 & 94 & 98 & 0.39 & 0.50 & 2.43 \\
\midrule
\multicolumn{13}{l}{\emph{Reference: the 50\% Phi-2 arm with seed 42}} \\
50\% Phi-2, 43 & 1.27 & 1.0 & 0.69 & 0.55 & 0.03 &  &  & 56 &  & $-$0.31 & 0.18 &  \\
25\% human, 42 & 2.38 & 1.9 & 2.32 & 0.39 & $-$0.33 & 1.63 & $-$0.16 & 86 & $\dagger$ & 0.20 & 0.50 & 0.70 \\
50\% human, 42 & 7.73 & 6.1 & 8.10 & 0.63 & $-$1.00 & 7.41 & 0.08 & 93 & 99 & 0.13 & 0.43 & 1.71 \\
\bottomrule
\end{tabular}
\end{table}

\subsection{What the members lose}
\label{sec:text}

In their text the members drift by shedding function words, and each keeps writing texts of its own. In the uniform arm the rate of common function words falls from 211 to 78 per thousand words over five generations, and it falls in every generation of every arm (Table~\ref{tab:fw}). The continuations lose articles and pronouns, their words grow longer, and late in a text they often run into chains of loosely related words (Tables~\ref{tab:rev_text} and~\ref{tab:excerpts}). An outside model reads the same change as text that grows harder to predict, since the mean GPT-2-large perplexity of the members other than the head rises in every arm at every generation (Table~\ref{tab:rev_ppl}). The members do not collapse onto a few texts. The type-token ratio within a text rises in every arm, and in the first five generations the distinct 4-gram fraction of every member in every arm stays at or above \slot{open.u4}{0.94}.

The two text readings give clocks that do not depend on the encoder. The function-word clock is coarse, since a second run of the uniform arm already reads 0.65 of a generation behind the first, and it does not resolve the 50\% arm, which reads 0.3 of a generation behind in one run and on time in the other (Table~\ref{tab:fw}). The perplexity clock is finer, with the two uniform runs 0.3 of a generation apart, and on it both runs of the 50\% arm fall behind by more than that (Table~\ref{tab:rev_ppl}). Both clocks put the 90\% arm behind every other arm without human text, as the encoder does.

\subsection{The members drift as one block}
\label{sec:direction}

\begin{figure}[t]
\centering
\includegraphics[width=0.75\textwidth]{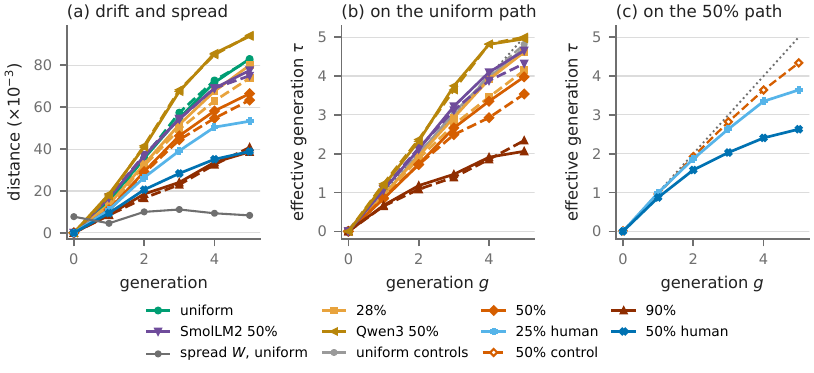}
\caption{Drift and effective generation by generation, one line per arm, with chain seed 42 solid and seed 43 dashed. (a)~Member drift $D^A(g)$ of each arm and the spread $W(g)$ of the uniform arm with seed 42 in grey. (b)~Effective generation $\tau^{B|\mathrm{uniform}}(g)$ of the 28\%, 50\%, 90\%, SmolLM2 and Qwen3 arms on the path of the uniform arm with the same chain seed. The grey controls place each uniform run on the other's path, seed 43 on seed 42 solid and seed 42 on seed 43 dashed. (c)~Effective generation of the two human arms on the path of the 50\% arm with seed 42, with its seed-43 run as control. The dotted diagonal is $\tau=g$.}
\label{fig:drift}
\end{figure}

The eight members travel together, far from where they started and never far from each other. In the uniform arm the spread between the members peaks at \slot{blk.smax}{11.2} over five generations while their mean drift reaches \slot{open.d5}{83} (Figure~\ref{fig:drift}a), and each member's drift points the way the other seven drift, with a co-movement cosine of \slot{blk.cm}{0.97} averaged over the members. Every other arm moves as a block in the same sense (Table~\ref{tab:rev_block}). The block also slows without turning, since from generation 5 to 10 the uniform arm adds only \slot{long.dd}{15} to its drift and the members keep their direction to a mean cosine of \slot{long.k}{0.99} (Table~\ref{tab:long}).

\begin{figure}[t]
\centering
\includegraphics[width=0.8\textwidth]{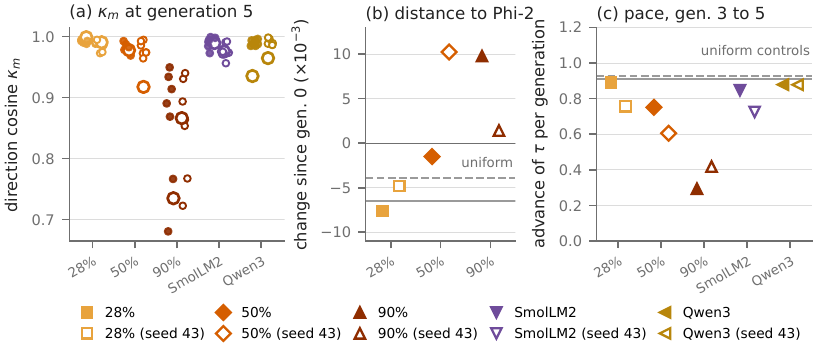}
\caption{Direction, distance to Phi-2 and pace, one column per arm, with seed 42 filled and seed 43 open. (a)~Direction cosine $\kappa_m$ between each member's drift in a concentrated arm and its drift in the uniform arm with the same chain seed at generation 5, one dot per member, with the head drawn as a large ring. (b)~Change from generation 0 to generation 5 of the mean distance from the seven other members to Phi-2 in the 28\%, 50\% and 90\% arms, with the two uniform runs as grey lines, seed 42 solid and seed 43 dashed. (c)~Mean advance of $\tau$ per generation over generations 3 to 5 on the uniform path of the same chain seed, with the two uniform controls, each uniform run placed on the other's path, as grey lines.}
\label{fig:direction}
\end{figure}

The direction cosine alone cannot say whether a supplier changes the route. Whether Phi-2 writes 28\% or half of the pool, or SmolLM2 or Qwen3 writes half, the members' drift points the same way as in the uniform arm to a mean cosine of at least \slot{abs.cos}{0.97} (Figure~\ref{fig:direction}a, Table~\ref{tab:rev_kappa}). Two runs of the uniform arm agree to \slot{kap.ctrl}{0.993}, and the half-human arm still agrees with the 50\% arm to \slot{kap.h50}{0.961}. Every arm shares one dominant drift axis, the loss of function words, so this cosine cannot tell a small change of route from none, and Section~\ref{sec:brake} splits the separation instead.

Nor does Phi-2 pull the other members toward it. If Phi-2 pulled the ecosystem, the other members would end closer to Phi-2 than in the uniform arm, and more so as its share grew. In the uniform arm their mean distance to Phi-2 falls by about half over five generations, and in the 50\% arm it falls less in one run and rises in the other, by about two and five times the difference between the two uniform runs (Figure~\ref{fig:direction}b, Table~\ref{tab:rev_headdist}). With 28\% the change stays within that difference, and in both runs at 90\% the members end farther from Phi-2 than in the uniform arm.

Two readings in this section depend on the encoder, the spread of the block and the distance to SmolLM2. Under all-mpnet-base-v2, a sentence encoder trained for semantic similarity, the spread of the uniform arm falls from 101 to 21 over five generations, the convergence that \citet{vu2025what} and \citet{wang2025llm} report, while under DeBERTa-v3 it stays flat. Under both encoders the members move together, with a co-movement cosine of 0.89 under all-mpnet-base-v2, the concentrated arms keep the same order of pace, and no arm brings the members closer to Phi-2 by more than the two uniform runs differ (Table~\ref{tab:encgeom}). Under all-mpnet-base-v2 the members of the SmolLM2 arm end slightly closer to SmolLM2 in one of its two runs, by a little more than that difference.

\subsection{Delay and departure}
\label{sec:brake}

The separation that Phi-2 adds beyond seed noise is mostly delay at 90\% and about half delay at half the pool. Two runs of the uniform arm differ almost entirely off the route, so a raw split counts their seed noise as departure. Beyond the second uniform run the delay makes up 54\% and 57\% of the two parts in the two runs of the 50\% arm, against 72\% and 64\% in the two runs at 90\% (Table~\ref{tab:split}). The two runs of the 50\% arm, by contrast, differ from each other more in delay than off the route, so at half the pool the ordering of the two parts lies within the arm's own seed variation. In one run the 28\% arm exceeds the second uniform run by less than one seed floor, too little for its excess to be split, and in the other by about two, of which delay makes up 43\%. With SmolLM2 as the head, the two runs disagree on whether the block separates at all, and where it separates, most of the excess lies off the route. The ordering of the two parts is a result about Phi-2.

The departure leans toward Phi-2 more as Phi-2's share grows, but up to half the pool it leans as far toward another member. Its cosine with the direction to Phi-2 rises from 0.36 and 0.54 at 28\% to 0.61 and 0.59 at 50\% and 0.72 and 0.73 at 90\% (Table~\ref{tab:split}). Taken toward each of the seven other members instead, the same cosine exceeds 0.5 for at least one of them in every arm without human text, the uniform controls included, and with Phi-2 or SmolLM2 as head that member is SmolLM3 (Table~\ref{tab:towardnull}). Phi-2 ties SmolLM3 at half the pool and exceeds it only at 90\%, in both runs. Any departure leans toward some member, and only the 90\% arm under DeBERTa-v3 singles out Phi-2.

Neither run of the 50\% arm brings the members closer to Phi-2, because the delay holds them back and, in one run, Phi-2 itself strays. Placing every member, Phi-2 included, at its nearest route point already puts the members of the 50\% arm farther from Phi-2 than in the uniform arm (Table~\ref{tab:rev_headdist}). The departures leave the members about there in the run with seed 42 and carry them 8.1 farther in the run with seed 43, where Phi-2 leaves the route by 10.6 against 2.0 (Table~\ref{tab:rev_offpath}). At half the pool the distance to the head follows where the head itself goes.

Measured in generations along the route, the 50\% arm falls further behind in every generation. On the path that each member traces in the uniform arm with the same chain seed, it trails a second run of the uniform arm at generation 5 by \slot{path.lag50}{0.75} and \slot{path.lag50s43}{1.3} generations in the two runs, and \slot{pm.pos}{15} of its 16 members across the two runs lag (Figure~\ref{fig:drift}b, Tables~\ref{tab:rev_offpath} and~\ref{tab:permember}).

With nine tenths of the pool, Phi-2 nearly stops drifting, and the block slows with it. As head, Phi-2 drifts 30 and 37 in the two runs against about 95 in the uniform arm, and the members as a whole drift less than half as far as in the uniform arm (Table~\ref{tab:members}). Through generation 4 Phi-2 never passes the point it reaches after one generation in the uniform arm, while it moves off that route (Table~\ref{tab:rev_offpath}). In the second run its last centroid lies nearly as close to an early point of the route as to a later one, so its position at generation 5 is ambiguous. From the third generation on the block advances less than half a generation per generation, against about 0.9 on average for the uniform controls (Figure~\ref{fig:direction}c), and after five generations it trails a second run of the uniform arm by \slot{path.lag90}{2.7} and 2.5 generations (Figure~\ref{fig:drift}b). The other members keep ahead of Phi-2 on the route, but at a fraction of the uniform pace.

\subsection{Who holds the head}
\label{sec:identity}

At half the pool, whether a head slows the block depends on which model it is. With SmolLM2 as head the block trails the second uniform run less than with Phi-2 at the same chain seed, in one run hardly at all (Table~\ref{tab:split}). With Qwen3-1.7B as head it runs ahead of the uniform arm on its path in every generation of both runs, by 0.7 of a generation at generation 3, before any member reaches the end of the path, and by 0.8 to 0.9 at generation 4, when four have (Figure~\ref{fig:drift}b). All-mpnet-base-v2 and the two text clocks do not separate it from the uniform arm by more than about one seed floor (Tables~\ref{tab:rev_encoder2}, \ref{tab:fw} and~\ref{tab:rev_ppl}). The members end farther from Qwen3 than in the uniform arm, and their departure leans no farther toward Qwen3 than toward another member, its sibling Qwen2.5-1.5B in one run and OLMo-2-1B in the other (Tables~\ref{tab:rev_headdist} and~\ref{tab:towardnull}). A half share can hold the block back or leave it as fast as equal shares.

The three heads slow the block in the order in which they slow themselves. As head, Phi-2 drifts a third to a half less than in the uniform arm, SmolLM2 about an eighth less and Qwen3 about as far or slightly farther, and the other seven members follow in the same order (Table~\ref{tab:members}). Perplexity rises along the route (Section~\ref{sec:text}), and as heads Phi-2 and SmolLM2 write more predictable text than the other members through generation 4, while Qwen3 writes less predictable text through generation 3 in one run and through generation 2 in the other (Table~\ref{tab:rev_ppl}). This is consistent with a head acting through its own pace.

\begin{figure}[t]
\centering
\includegraphics[width=0.75\textwidth]{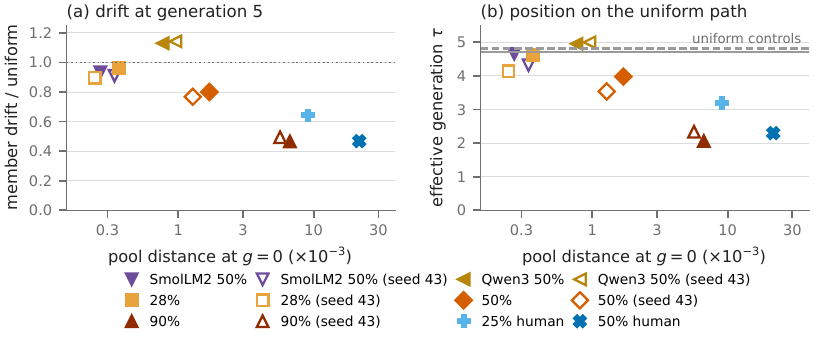}
\caption{Slowdown against the distance between an arm's generation-0 pool centroid and the uniform arm's, on a logarithmic axis, one marker per arm and chain seed, with seed 43 open. The human pools include their human text, and both human arms are read on the uniform path with seed 42. (a)~Member drift $D$ at generation 5 over that of the uniform arm with the same chain seed. (b)~Effective generation $\tau$ at generation 5 on the path of the uniform arm with the same chain seed, with the uniform controls in grey as in Figure~\ref{fig:drift}b.}
\label{fig:pull}
\end{figure}

How far a share moves the pool sets how much a head can change the pace. At generation 0 the pool of the 50\% arm lies \slot{pull.p50}{1.7} and \slot{pull.p50s43}{1.3} from the uniform pool in the two runs, against \slot{pull.pS}{0.27} and 0.34 for the pool of the SmolLM2 arm (Table~\ref{tab:rev_pool}). Across the eight runs with Phi-2 or SmolLM2 as head, the slowdown rises with this distance in three groups under all-mpnet-base-v2 and, less cleanly, under DeBERTa-v3, from the 28\% and SmolLM2 arms through the 50\% arm to the 90\% arm (Figure~\ref{fig:pull}, Table~\ref{tab:rev_encoder2}). The pool of the Qwen3 arm lies farther from the uniform pool than those of the 28\% and SmolLM2 arms, yet its block does not slow, which is consistent with the distance setting how much a head changes the pace and the head's own pace setting which way.

\subsection{Human text}
\label{sec:human}

Human text slows the members along the route they already follow, and more so the larger its part of the pool. With half the pool human the members drift \slot{hum.r50}{42}\% less than in the 50\% arm after five generations and trail its second run along its path by \slot{hum.lag50}{1.7} generations (Figure~\ref{fig:drift}c), and almost all of their separation from the 50\% arm beyond that run lies along the path (Table~\ref{tab:split}). With a quarter of the pool human they drift \slot{hum.r25}{20}\% less and trail that run by \slot{hum.lag25}{0.7} of a generation, without straying farther from the path than it does. Human text thus acts almost only on the members' pace.

Human text slows the members less than its distance from their text would predict. Placed on Figure~\ref{fig:pull}, both human pools lie farther from the uniform pool than the pool of the 90\% arm does, the half-human pool more than three times as far, yet on the uniform path the half-human arm trails only about as far as the 90\% arm and the quarter-human arm less far (Tables~\ref{tab:rev_pool} and~\ref{tab:split}). One reading is that only the part of a pool's difference that lies along the members' route slows them.

\section{Discussion and conclusion}
\label{sec:discussion}

The worry about concentration pictures a dominant supplier whose text pulls the other models toward the way it writes. In our ecosystems no share of Phi-2 pulls the others closer to it, and their departures from the common route single it out only at nine tenths of the pool, while at half the pool or more it slows the whole block. Every member, the dominant one included, travels along one common route, and a supplier that writes much of the pool and drifts slowly itself fills it with text from an earlier point on that route, which can hold the others back. At half the pool Phi-2, which moves the pool far and slows itself, brakes the block, and Qwen3-1.7B, which keeps its own pace, does not. Human text, drawn afresh from a fixed corpus, also slows the block along its route.

The route itself comes from the generation recipe that all members share. Every member samples with a repetition penalty, which lowers the score of every token the continuation has already used, and a frequency penalty, which lowers it further with each use, and function words are the tokens a sentence repeats most. The recipe also cuts every continuation at 128 new tokens, and nearly nine in ten texts already end mid-sentence at generation 0 (Table~\ref{tab:rev_text}), so training on unfinished text could drive the same drift. A variant of the recipe that drops both penalties, and keeps the cut and a filter that discards repetitive texts, separates the two (Table~\ref{tab:recipe}). Without the penalties the base models write 302 function words per thousand instead of 211, and the rate rises in each of the next four generations, so the loss of function words comes from the penalties. The members also start repeating themselves, and by generation 4 the filter rejects about half of the uniform arm's continuations. Along this new route a half share of Phi-2 still slows the block, which trails the variant's uniform arm by 1.1 generations at generation 4, against 0.6 and 1.1 for the 50\% arm under the main recipe. In this variant the recipe set the route, and Phi-2's share still set the pace along it.

Other multi-model studies find that models converge, and under the sentence encoder our members converge too, though not on the supplier. \citet{vu2025what} continue training each model from its previous weights for fifteen generations, and in \citet{wang2025llm} the models share a retrieval memory, so in both a model's exposure to the others builds up. Our members restart from their base weights in every generation, so a supplier's influence cannot build up in the weights, and they converge on a common style without being drawn to whoever writes most. Our reading is that convergence on a dominant model needs a coupling that accumulates.

\citet{hodel2025epistemic} find that concentration speeds collapse, and the difference lies in the variable. In their setup a fixed corpus is divided among models that each train on their own part, so concentrating it in fewer models removes the diversity across models that protects the population. Here every member trains on the whole pool, and concentration changes only what the pool is made of. \citet{wu2026when} show that which models collapse is set by the structure of the graph of who trains on whose output. All our arms share one graph, in which every member trains on text from every member, and the shares that vary between them change the pace of drift.

\paragraph{Limitations.} The human arms have one chain seed each and are read against the seed floor of the uniform arm, and by generation 5 seven or eight members of the Qwen3 arm have reached the end of the uniform path, which bounds its lead only from below. The reading of the pace through pool distance and the head's own pace rests on ten concentrated runs with three heads, among which a head's own pace and the predictability of its text go together. The coupling is light, one epoch from the base weights on 2{,}100 texts of at most 128 generated tokens, and a heavier dose or training that continues from the previous weights might let a supplier pull the members or make them converge on it. The steering depends on the encoder, and under all-mpnet-base-v2 no departure leans toward Phi-2 more than toward every other member. The recipe variant has one run of four generations. The ecosystem has eight members of 1B to 3B parameters, runs five generations, ten for the uniform arm, and draws its human text from one corpus.

Recursive training among several models has been pictured as a convergence on whoever writes most. In our ecosystems the members converge on a common route that the recipe sets, no share of Phi-2 pulls them closer to it, and a dominant supplier changes the drift mostly through its pace: Phi-2 at nine tenths of the pool holds the block about two and a half generations behind, and which model writes half decides whether the block slows at all.

\clearpage
\subsubsection*{Ethics statement}
This work trains and evaluates only publicly released open-weight models on public research corpora. No human subjects, personal data or annotators are involved. The human text is a filtered subset of pile-10k, a public sample of the Pile whose Books3 component contains copyrighted books. It is used only as training data inside the simulation, and the paper redistributes no corpus text beyond the two short prompts and the excerpts of model output in Table~\ref{tab:excerpts}. The findings bear on discussions of data governance but make no policy recommendation.

\subsubsection*{Reproducibility statement}
All models, corpora and scoring instruments are public. Appendix~\ref{app:methods} gives the roster with full identifiers, the shares and quotas, the per-generation pipeline, the training and generation recipes, the seed assignment, the encoder, the text statistics and the human-corpus filters. The analysis scripts compute every number in the paper from the stored member centroids and generated texts, and the table and figure scripts read their output. Experiment code, configurations and the member centroids will be released.

\subsubsection*{AI use statement}
We used generative AI tools to implement standard components of the experimental pipeline, to assist with data analysis and figure drawing, and to draft and edit parts of the paper. The recursively generated text that this paper studies is produced by the open-weight models under examination, following the procedures of Section~\ref{sec:setup} and Appendix~\ref{app:methods}. The authors read and tested the pipeline code before any reported run and read and approved all drafted and edited text. We take responsibility for the final content of this work, including text, claims and artifacts produced with the aid of generative AI.

\bibliography{references}

\begin{thebibliography}{40}
\providecommand{\natexlab}[1]{#1}
\providecommand{\url}[1]{\texttt{#1}}
\expandafter\ifx\csname urlstyle\endcsname\relax
  \providecommand{\doi}[1]{doi: #1}\else
  \providecommand{\doi}{doi: \begingroup \urlstyle{rm}\Url}\fi

\bibitem[Alemohammad et~al.(2024)Alemohammad, Casco-Rodriguez, Luzi, Humayun,
  Babaei, LeJeune, Siahkoohi, and Baraniuk]{alemohammad2024selfconsuming}
Sina Alemohammad, Josue Casco-Rodriguez, Lorenzo Luzi, Ahmed~Imtiaz Humayun,
  Hossein Babaei, Daniel LeJeune, Ali Siahkoohi, and Richard~G. Baraniuk.
\newblock Self-consuming generative models go {MAD}.
\newblock In \emph{International Conference on Learning Representations}, 2024.
\newblock arXiv:2307.01850.

\bibitem[Bellagente et~al.(2024)Bellagente, Tow, Mahan, Phung, Zhuravinskyi,
  Adithyan, Baicoianu, Brooks, Cooper, Datta, Lee, Mostaque, Pieler,
  Pinnaparju, Rocha, Saini, Teufel, Zanichelli, and
  Riquelme]{bellagente2024stable}
Marco Bellagente, Jonathan Tow, Dakota Mahan, Duy Phung, Maksym Zhuravinskyi,
  Reshinth Adithyan, James Baicoianu, Ben Brooks, Nathan Cooper, Ashish Datta,
  Meng Lee, Emad Mostaque, Michael Pieler, Nikhil Pinnaparju, Paulo Rocha,
  Harry Saini, Hannah Teufel, Niccolo Zanichelli, and Carlos Riquelme.
\newblock Stable {LM} 2 1.6{B} technical report, 2024.
\newblock arXiv:2402.17834.

\bibitem[{Ben Allal} et~al.(2025){Ben Allal}, Lozhkov, Bakouch, Bl{\'a}zquez,
  Penedo, Tunstall, Marafioti, Kydl{\'i}{\v{c}}ek, Lajar{\'i}n, Srivastav,
  Lochner, Fahlgren, Nguyen, Fourrier, Burtenshaw, Larcher, Zhao, Zakka,
  Morlon, Raffel, von Werra, and Wolf]{benallal2025smollm2}
Loubna {Ben Allal}, Anton Lozhkov, Elie Bakouch, Gabriel~Mart{\'i}n
  Bl{\'a}zquez, Guilherme Penedo, Lewis Tunstall, Andr{\'e}s Marafioti, Hynek
  Kydl{\'i}{\v{c}}ek, Agust{\'i}n~Piqueres Lajar{\'i}n, Vaibhav Srivastav,
  Joshua Lochner, Caleb Fahlgren, Xuan-Son Nguyen, Cl{\'e}mentine Fourrier, Ben
  Burtenshaw, Hugo Larcher, Haojun Zhao, Cyril Zakka, Mathieu Morlon, Colin
  Raffel, Leandro von Werra, and Thomas Wolf.
\newblock {SmolLM2}: {When} smol goes big -- data-centric training of a small
  language model, 2025.
\newblock arXiv:2502.02737.

\bibitem[Bertrand et~al.(2024)Bertrand, Bose, Duplessis, Jiralerspong, and
  Gidel]{bertrand2024stability}
Quentin Bertrand, Avishek~Joey Bose, Alexandre Duplessis, Marco Jiralerspong,
  and Gauthier Gidel.
\newblock On the stability of iterative retraining of generative models on
  their own data.
\newblock In \emph{International Conference on Learning Representations}, 2024.
\newblock arXiv:2310.00429.

\bibitem[Bommasani et~al.(2022)Bommasani, Creel, Kumar, Jurafsky, and
  Liang]{bommasani2022picking}
Rishi Bommasani, Kathleen~A. Creel, Ananya Kumar, Dan Jurafsky, and Percy
  Liang.
\newblock Picking on the same person: Does algorithmic monoculture lead to
  outcome homogenization?
\newblock In \emph{Advances in Neural Information Processing Systems}, 2022.
\newblock arXiv:2211.13972.

\bibitem[Briesch et~al.(2023)Briesch, Sobania, and Rothlauf]{briesch2023large}
Martin Briesch, Dominik Sobania, and Franz Rothlauf.
\newblock Large language models suffer from their own output: An analysis of
  the self-consuming training loop, 2023.
\newblock arXiv:2311.16822.

\bibitem[{Competition and Markets Authority}(2024)]{cma2024ai}
{Competition and Markets Authority}.
\newblock {AI} foundation models: Update paper.
\newblock UK Competition and Markets Authority, April 2024.
\newblock URL
  \url{https://www.gov.uk/government/publications/ai-foundation-models-update-paper}.

\bibitem[Dohmatob et~al.(2024)Dohmatob, Feng, Yang, Charton, and
  Kempe]{dohmatob2024tale}
Elvis Dohmatob, Yunzhen Feng, Pu~Yang, Fran{\c c}ois Charton, and Julia Kempe.
\newblock A tale of tails: {Model} collapse as a change of scaling laws.
\newblock In \emph{International Conference on Machine Learning}, 2024.
\newblock arXiv:2402.07043.

\bibitem[{Falcon-LLM Team}(2024)]{falcon2024falcon3}
{Falcon-LLM Team}.
\newblock The falcon 3 family of open models.
\newblock \url{https://huggingface.co/blog/falcon3}, December 2024.

\bibitem[Feng et~al.(2025)Feng, Dohmatob, Yang, Charton, and
  Kempe]{feng2024beyond}
Yunzhen Feng, Elvis Dohmatob, Pu~Yang, Fran{\c c}ois Charton, and Julia Kempe.
\newblock Beyond model collapse: Scaling up with synthesized data requires
  verification.
\newblock In \emph{International Conference on Learning Representations}, 2025.
\newblock arXiv:2406.07515.

\bibitem[Gao et~al.(2020)Gao, Biderman, Black, Golding, Hoppe, Foster, Phang,
  He, Thite, Nabeshima, Presser, and Leahy]{gao2020pile}
Leo Gao, Stella Biderman, Sid Black, Laurence Golding, Travis Hoppe, Charles
  Foster, Jason Phang, Horace He, Anish Thite, Noa Nabeshima, Shawn Presser,
  and Connor Leahy.
\newblock The {Pile}: {An} 800{GB} dataset of diverse text for language
  modeling, 2020.
\newblock arXiv:2101.00027.

\bibitem[Gerstgrasser et~al.(2024)Gerstgrasser, Schaeffer, Dey, Rafailov,
  Korbak, Sleight, Agrawal, Hughes, Pai, Gromov, Roberts, Yang, Donoho, and
  Koyejo]{gerstgrasser2024model}
Matthias Gerstgrasser, Rylan Schaeffer, Apratim Dey, Rafael Rafailov, Tomasz
  Korbak, Henry Sleight, Rajashree Agrawal, John Hughes, Dhruv~Bhandarkar Pai,
  Andrey Gromov, Dan Roberts, Diyi Yang, David~L. Donoho, and Sanmi Koyejo.
\newblock Is model collapse inevitable? {Breaking} the curse of recursion by
  accumulating real and synthetic data.
\newblock In \emph{First Conference on Language Modeling (COLM)}, 2024.
\newblock arXiv:2404.01413.

\bibitem[Gillman et~al.(2024)Gillman, Freeman, Aggarwal, Hsu, Luo, Tian, and
  Sun]{gillman2024selfcorrecting}
Nate Gillman, Michael Freeman, Daksh Aggarwal, Chia-Hong Hsu, Calvin Luo,
  Yonglong Tian, and Chen Sun.
\newblock Self-correcting self-consuming loops for generative model training.
\newblock In \emph{International Conference on Machine Learning}, 2024.
\newblock arXiv:2402.07087.

\bibitem[Guo et~al.(2024)Guo, Shang, Vazirgiannis, and Clavel]{guo2024curious}
Yanzhu Guo, Guokan Shang, Michalis Vazirgiannis, and Chlo{\'e} Clavel.
\newblock The curious decline of linguistic diversity: Training language models
  on synthetic text.
\newblock In \emph{Findings of the Association for Computational Linguistics:
  NAACL 2024}, pp.\  3589--3604, 2024.
\newblock arXiv:2311.09807.

\bibitem[He et~al.(2023)He, Gao, and Chen]{he2023debertav3}
Pengcheng He, Jianfeng Gao, and Weizhu Chen.
\newblock {DeBERTaV3}: {Improving} {DeBERTa} using {ELECTRA}-style pre-training
  with gradient-disentangled embedding sharing.
\newblock In \emph{International Conference on Learning Representations}, 2023.
\newblock arXiv:2111.09543.

\bibitem[Hodel \& West(2025)Hodel and West]{hodel2025epistemic}
Damian Hodel and Jevin~D. West.
\newblock Epistemic diversity across language models mitigates knowledge
  collapse, 2025.
\newblock arXiv:2512.15011.

\bibitem[Holtzman et~al.(2020)Holtzman, Buys, Du, Forbes, and
  Choi]{holtzman2020curious}
Ari Holtzman, Jan Buys, Li~Du, Maxwell Forbes, and Yejin Choi.
\newblock The curious case of neural text degeneration.
\newblock In \emph{International Conference on Learning Representations}, 2020.
\newblock arXiv:1904.09751.

\bibitem[{Hugging Face SmolLM Team}(2025)]{bakouch2025smollm3}
{Hugging Face SmolLM Team}.
\newblock {SmolLM3}: smol, multilingual, long-context reasoner.
\newblock \url{https://huggingface.co/blog/smollm3}; model card,
  \url{https://huggingface.co/HuggingFaceTB/SmolLM3-3B-Base}, 2025.

\bibitem[Kazdan et~al.(2025)Kazdan, Schaeffer, Dey, Gerstgrasser, Rafailov,
  Donoho, and Koyejo]{kazdan2024collapse}
Joshua Kazdan, Rylan Schaeffer, Apratim Dey, Matthias Gerstgrasser, Rafael
  Rafailov, David~L. Donoho, and Sanmi Koyejo.
\newblock Collapse or thrive: {Perils} and promises of synthetic data in a
  self-generating world.
\newblock In \emph{Proceedings of the 42nd International Conference on Machine
  Learning}, volume 267 of \emph{Proceedings of Machine Learning Research},
  pp.\  29469--29494, 2025.
\newblock arXiv:2410.16713.

\bibitem[Kleinberg \& Raghavan(2021)Kleinberg and
  Raghavan]{kleinberg2021algorithmic}
Jon Kleinberg and Manish Raghavan.
\newblock Algorithmic monoculture and social welfare.
\newblock \emph{Proceedings of the National Academy of Sciences}, 118\penalty0
  (22):\penalty0 e2018340118, 2021.

\bibitem[Korinek \& Vipra(2025)Korinek and Vipra]{korinek2025concentrating}
Anton Korinek and Jai Vipra.
\newblock Concentrating intelligence: scaling and market structure in
  artificial intelligence.
\newblock \emph{Economic Policy}, 40\penalty0 (121):\penalty0 225--256, 2025.
\newblock \doi{10.1093/epolic/eiae057}.

\bibitem[Kwon et~al.(2023)Kwon, Li, Zhuang, Sheng, Zheng, Yu, Gonzalez, Zhang,
  and Stoica]{kwon2023efficient}
Woosuk Kwon, Zhuohan Li, Siyuan Zhuang, Ying Sheng, Lianmin Zheng, Cody~Hao Yu,
  Joseph~E. Gonzalez, Hao Zhang, and Ion Stoica.
\newblock Efficient memory management for large language model serving with
  {PagedAttention}.
\newblock In \emph{Proceedings of the 29th Symposium on Operating Systems
  Principles (SOSP)}, 2023.
\newblock arXiv:2309.06180.

\bibitem[Laurer et~al.(2024)Laurer, van Atteveldt, Casas, and
  Welbers]{laurer2022less}
Moritz Laurer, Wouter van Atteveldt, Andreu Casas, and Kasper Welbers.
\newblock Less annotating, more classifying: Addressing the data scarcity issue
  of supervised machine learning with deep transfer learning and {BERT-NLI}.
\newblock \emph{Political Analysis}, 32\penalty0 (1):\penalty0 84--100, 2024.
\newblock \doi{10.1017/pan.2023.20}.

\bibitem[Law(2025)]{law2025ai}
Ryan Law.
\newblock 74\% of new webpages include {AI} content (study of 900k pages).
\newblock Ahrefs Blog, May 2025.
\newblock URL
  \url{https://ahrefs.com/blog/what-percentage-of-new-content-is-ai-generated/}.

\bibitem[Li et~al.(2016)Li, Galley, Brockett, Gao, and Dolan]{li2016diversity}
Jiwei Li, Michel Galley, Chris Brockett, Jianfeng Gao, and Bill Dolan.
\newblock A diversity-promoting objective function for neural conversation
  models.
\newblock In \emph{Proceedings of the 2016 Conference of the North American
  Chapter of the Association for Computational Linguistics: Human Language
  Technologies (NAACL-HLT)}, 2016.
\newblock arXiv:1510.03055.

\bibitem[{Microsoft Research}(2023)]{microsoft2023phi2}
{Microsoft Research}.
\newblock Phi-2: {The} surprising power of small language models.
\newblock Model card, \url{https://huggingface.co/microsoft/phi-2};
  announcement,
  \url{https://www.microsoft.com/en-us/research/blog/phi-2-the-surprising-power-of-small-language-models/},
  2023.

\bibitem[Nanda(2022)]{nanda2022pile10k}
Neel Nanda.
\newblock pile-10k.
\newblock Dataset, \url{https://huggingface.co/datasets/NeelNanda/pile-10k},
  2022.
\newblock The first 10{,}000 documents of the Pile (Gao et al., 2020).

\bibitem[Radford et~al.(2019)Radford, Wu, Child, Luan, Amodei, and
  Sutskever]{radford2019language}
Alec Radford, Jeffrey Wu, Rewon Child, David Luan, Dario Amodei, and Ilya
  Sutskever.
\newblock Language models are unsupervised multitask learners.
\newblock \emph{OpenAI blog}, 1\penalty0 (8), 2019.

\bibitem[Schaeffer et~al.(2025)Schaeffer, Kazdan, Arulandu, and
  Koyejo]{schaeffer2025position}
Rylan Schaeffer, Joshua Kazdan, Alvan~Caleb Arulandu, and Sanmi Koyejo.
\newblock Position: Model collapse does not mean what you think, 2025.
\newblock arXiv:2503.03150.

\bibitem[Seddik et~al.(2024)Seddik, Chen, Hayou, Youssef, and
  Debbah]{seddik2024bad}
Mohamed El~Amine Seddik, Suei-Wen Chen, Soufiane Hayou, Pierre Youssef, and
  Merouane Debbah.
\newblock How bad is training on synthetic data? {A} statistical analysis of
  language model collapse.
\newblock In \emph{First Conference on Language Modeling (COLM)}, 2024.
\newblock arXiv:2404.05090.

\bibitem[Shumailov et~al.(2024)Shumailov, Shumaylov, Zhao, Papernot, Anderson,
  and Gal]{shumailov2024ai}
Ilia Shumailov, Zakhar Shumaylov, Yiren Zhao, Nicolas Papernot, Ross Anderson,
  and Yarin Gal.
\newblock {AI} models collapse when trained on recursively generated data.
\newblock \emph{Nature}, 631\penalty0 (8022):\penalty0 755--759, 2024.
\newblock \doi{10.1038/s41586-024-07566-y}.
\newblock arXiv:2305.17493.

\bibitem[{Team OLMo} et~al.(2024){Team OLMo}, Walsh, Soldaini, Groeneveld, Lo,
  Arora, Bhagia, Gu, Huang, Jordan, Lambert, Schwenk, Tafjord, Anderson,
  Atkinson, Brahman, Clark, Dasigi, Dziri, Ettinger, et~al.]{teamolmo2024olmo2}
{Team OLMo}, Pete Walsh, Luca Soldaini, Dirk Groeneveld, Kyle Lo, Shane Arora,
  Akshita Bhagia, Yuling Gu, Shengyi Huang, Matt Jordan, Nathan Lambert, Dustin
  Schwenk, Oyvind Tafjord, Taira Anderson, David Atkinson, Faeze Brahman,
  Christopher Clark, Pradeep Dasigi, Nouha Dziri, Allyson Ettinger, et~al.
\newblock 2 {OLMo} 2 furious, 2024.
\newblock arXiv:2501.00656.

\bibitem[Tully et~al.(2025)Tully, Redfern, Das, and Xiao]{menlo2025state}
Tim Tully, Joff Redfern, Deedy Das, and Derek Xiao.
\newblock 2025: The state of generative {AI} in the enterprise.
\newblock Menlo Ventures, December 2025.
\newblock URL
  \url{https://menlovc.com/perspective/2025-the-state-of-generative-ai-in-the-enterprise/}.

\bibitem[Turegeldinova et~al.(2025)Turegeldinova, Amralinova, Fodor, Eraliyeva,
  Dayou, and Joldassov]{turegeldinova2025ai}
Aliya Turegeldinova, Bakytzhan Amralinova, Mate~Miklos Fodor, Akerkin
  Eraliyeva, Chen Dayou, and Aidos Joldassov.
\newblock {AI} as a centripetal technology: {Price} compression,
  homogenization, and entry, 2025.
\newblock arXiv:2510.08337.

\bibitem[Vu et~al.(2025)Vu, Reeves, and Wenger]{vu2025what}
Hung~Anh Vu, Galen Reeves, and Emily Wenger.
\newblock What happens when generative {AI} models train recursively on each
  others' outputs?, 2025.
\newblock arXiv:2505.21677.

\bibitem[Wang et~al.(2025)Wang, Horiguchi, Pang, and Priebe]{wang2025llm}
Tianyu Wang, Akira Horiguchi, Lingyou Pang, and Carey~E. Priebe.
\newblock {LLM} web dynamics: {Tracing} model collapse in a network of {LLMs},
  2025.
\newblock arXiv:2506.15690.

\bibitem[Wu et~al.(2025)Wu, Black, and Chandrasekaran]{wu2024generative}
Fan Wu, Emily Black, and Varun Chandrasekaran.
\newblock Generative monoculture in large language models.
\newblock In \emph{International Conference on Learning Representations}, 2025.
\newblock arXiv:2407.02209.

\bibitem[Wu et~al.(2026)Wu, Zhou, and Su]{wu2026when}
Yuchen Wu, Kangjie Zhou, and Weijie Su.
\newblock When does model collapse occur in structured interactive learning?,
  2026.
\newblock arXiv:2605.20151.

\bibitem[Yang et~al.(2024)Yang, Yang, Zhang, Hui, Zheng, Yu, Li, Liu, Huang,
  et~al.]{qwen2024qwen25}
An~Yang, Baosong Yang, Beichen Zhang, Binyuan Hui, Bo~Zheng, Bowen Yu,
  Chengyuan Li, Dayiheng Liu, Fei Huang, et~al.
\newblock Qwen2.5 technical report, 2024.
\newblock arXiv:2412.15115.

\bibitem[Yang et~al.(2025)Yang, Li, Yang, Zhang, Hui, Zheng, Yu, Gao, Huang,
  Lv, Zheng, Liu, Zhou, Huang, Hu, Ge, Wei, Lin, Tang, Yang, Tu, Zhang, Yang,
  Yang, Zhou, Zhou, Lin, et~al.]{yang2025qwen3}
An~Yang, Anfeng Li, Baosong Yang, Beichen Zhang, Binyuan Hui, Bo~Zheng, Bowen
  Yu, Chang Gao, Chengen Huang, Chenxu Lv, Chujie Zheng, Dayiheng Liu, Fan
  Zhou, Fei Huang, Feng Hu, Hao Ge, Haoran Wei, Huan Lin, Jialong Tang, Jian
  Yang, Jianhong Tu, Jianwei Zhang, Jianxin Yang, Jiaxi Yang, Jing Zhou,
  Jingren Zhou, Junyang Lin, et~al.
\newblock Qwen3 technical report, 2025.
\newblock arXiv:2505.09388.

\end{thebibliography}
\bibliographystyle{iclr2027_conference}

\clearpage
\appendix
\makeatletter
\let\p@origtable\table
\let\p@origendtable\endtable
\renewenvironment{table}[1][]{\p@origtable[!htbp]}{\p@origendtable}
\makeatother
\renewcommand{\topfraction}{0.95}
\renewcommand{\bottomfraction}{0.9}
\renewcommand{\textfraction}{0.05}
\renewcommand{\floatpagefraction}{0.75}
\setcounter{topnumber}{4}
\setcounter{bottomnumber}{3}
\setcounter{totalnumber}{6}

\section{Methods}
\label{app:methods}

\paragraph{Members.} Table~\ref{tab:models} lists the eight members with their full identifiers. The last two columns give each member's mean distance, at generation 0, to the other seven members' centroids under each chain seed. Phi-2 has the largest value under both chain seeds, which is why it serves as the head.

\begin{table}[t]
\caption{The eight members, whose technical reports are, in row order, \citet{yang2025qwen3,qwen2024qwen25,benallal2025smollm2,bakouch2025smollm3,teamolmo2024olmo2,bellagente2024stable,microsoft2023phi2,falcon2024falcon3}. The last two columns give each member's mean cosine distance, in units of $10^{-3}$, at generation 0 to the other seven members' centroids, under chain seeds 42 and 43.}
\label{tab:models}
\centering
\small
\setlength{\tabcolsep}{4pt}
\begin{tabular}{llllrr}
\toprule
Member & Identifier & Size & Organization & Seed 42 & Seed 43 \\
\midrule
Qwen3-1.7B & \texttt{Qwen/Qwen3-1.7B-Base} & 1.7B & Alibaba & 10.9 & 9.5 \\
Qwen2.5-1.5B & \texttt{Qwen/Qwen2.5-1.5B} & 1.5B & Alibaba & 9.5 & 9.0 \\
SmolLM2-1.7B & \texttt{HuggingFaceTB/SmolLM2-1.7B} & 1.7B & HuggingFace & 5.7 & 4.9 \\
SmolLM3-3B & \texttt{HuggingFaceTB/SmolLM3-3B-Base} & 3B & HuggingFace & 4.4 & 4.2 \\
OLMo-2-1B & \texttt{allenai/OLMo-2-0425-1B} & 1B & AI2 & 6.8 & 6.7 \\
StableLM-2-1.6B & \texttt{stabilityai/stablelm-2-1\_6b} & 1.6B & Stability & 5.0 & 5.1 \\
Phi-2 & \texttt{microsoft/phi-2} & 2.7B & Microsoft & 13.5 & 12.1 \\
Falcon3-1B & \texttt{tiiuae/Falcon3-1B-Base} & 1B & TII & 6.2 & 6.0 \\
\bottomrule
\end{tabular}
\end{table}

\paragraph{Shares and quotas.} The pool always holds 2{,}100 texts. The uniform arm takes 262 or 263 texts from each member, and the concentrated arms take the head's share of 2{,}100 from the head and split the remainder evenly among the other seven, with largest-remainder rounding keeping the total fixed. In the 28\% arm the head supplies 588 texts and every other member 216. In the 50\% arm the head supplies 1{,}050 and every other member 150, and in the 90\% arm the head supplies 1{,}890 and every other member 30. The SmolLM2 and Qwen3 arms have the shares of the 50\% arm with SmolLM2 or Qwen3-1.7B as the head. The human arms keep the 50\% arm's proportions among the members and fill 525 or 1{,}050 of the 2{,}100 places with human excerpts, so that Phi-2 supplies 37.5\% or 25\% of the pool. Each member writes 800 texts per generation, and a head whose quota exceeds 800 writes its full quota. At generation 0 every member continues the same 800 prompts with its base weights, and a head's additional texts continue prompts 801 to 2{,}100 of the same prompt file. Pools are drawn from each member's texts without replacement.

\paragraph{Generation.} Texts are generated with vLLM \citep{kwon2023efficient} at temperature 1.0, top-p 0.95 \citep{holtzman2020curious}, repetition penalty 1.15, frequency penalty 0.3 and at most 128 new tokens. From generation 1 on, each member generates twice as many continuations as it needs (2.6 times for a head), and a filter that does not look at content keeps texts with at least 20 words, a most frequent word of at most 20\% of the words and a distinct-2 fraction \citep{li2016diversity} of at least 0.45, after which the texts are trimmed to the quota. Prompts are 8 to 16 words cut from human text. The model sees the prompt and its continuation during fine-tuning, and the encoder sees only the continuation.

\paragraph{Recipe variant.} A variant of the generation recipe sets the repetition penalty to 1.0 and the frequency penalty to 0 and keeps every other setting, the 128-token cut and the filter included. Because the filter rejects more of its output, each member draws eight times as many continuations as it needs. The variant has its own generation 0, written by the base models under it, and one run each of the uniform and 50\% arms. Its uniform arm ends after generation 4, when fewer than 800 of Qwen3-1.7B's 6{,}400 continuations pass the filter, so every reading of the variant stops there. The pass rate of a member is the fraction of the continuations the filter reads before it has filled the quota that pass it. Effective generations place the 50\% arm on the path of the variant's own uniform arm, and because the variant has one uniform run, the lag and the parts beyond the control use the main recipe's control at the same generation (Table~\ref{tab:recipe}).

\paragraph{Sampling seeds.} Request $i$ of member $k$ at generation $g$ under chain seed $s$ has seed $k\cdot 10^{8}+s\cdot 10^{6}+g\cdot 10^{5}+i$. Arms of one chain seed share generation 0 and share prompts and seeds at every later generation, and the two chain seeds share the generation-0 prompts but no seed.

\paragraph{Training.} Every member is fine-tuned from its base weights on the full pool of its generation, with all parameters updated, for one epoch at learning rate $2\times10^{-5}$ with 40 warmup steps, weight decay 0.01, an effective batch of 16 sequences, bf16 precision and a maximum sequence length of 768 tokens.

\paragraph{Human text.} Excerpts come from pile-10k \citep{nanda2022pile10k}, a 10{,}000-document sample of the Pile \citep{gao2020pile}. We keep the Pile-CC, OpenWebText2, Books3, BookCorpus2 and Gutenberg components and exclude arXiv, Wikipedia, PubMed, code and mathematics. Each excerpt has at least 50 words, no document contributes more than six excerpts, and each generation draws new excerpts.

\paragraph{Encoders and centroids.} The encoder is \texttt{MoritzLaurer/\allowbreak DeBERTa-\allowbreak v3-\allowbreak base-\allowbreak mnli-\allowbreak fever-\allowbreak anli}, never trained further. Each text is truncated to 128 tokens, mean-pooled over its attention mask and normalized to unit length, and a centroid is the normalized mean of these vectors. A member's centroid uses the first 800 texts it writes in a generation, including for heads that write more. A pool centroid uses all 2{,}100 pool texts. Since $d(x,y)=\tfrac12\lVert x-y\rVert^2$ for unit vectors, a separation divided by drift is the square of a ratio of root-mean-square chord lengths, and Table~\ref{tab:arms} gives its square root, the chord ratio, alongside it. The second encoder, \path{sentence-transformers/all-mpnet-base-v2}, embeds the same 800 texts per member, and Table~\ref{tab:rev_encoder2} compares the two.

\paragraph{Text statistics and perplexity.} Word counts use the texts that enter a member's centroid. Function words and type-token ratios count lowercase word tokens, and distinct 4-grams count whitespace-separated tokens pooled over a member's texts. The function-word rate of a member pools its texts, and an arm's rate averages its eight members. Perplexity is computed per token under GPT-2-large from the first 400 of the 800 texts, each truncated to 128 tokens, and its clock places the mean over the seven members other than the head on the uniform arm's curve for the same seven members, as in Table~\ref{tab:rev_ppl}.

\paragraph{Effective generation.} Each segment of the path $p^A_m$ between consecutive generations is sampled at 201 evenly spaced positions and renormalized to unit length, and the position with the smallest distance to $c^B_{m,g}$ gives $t_m(g)$ as the segment index plus the fraction along it. The control for an arm with chain seed 42 places the uniform run with seed 43 on the seed-42 uniform path, the control for an arm with seed 43 places the uniform run with seed 42 on the seed-43 path, and the control for the human arm places the 50\% run with seed 43 on the path of the 50\% run with seed 42. The departure cosine toward a member $j$ averages, over the seven members other than $j$, the cosine between a member's departure and the vector from its nearest route point to $c^B_{j,g}$, and Table~\ref{tab:towardnull} gives it for every $j$. The pace-matched distance to the head places every member, the head included, at its nearest point on the uniform path of the same chain seed and averages the distances from the seven other members to the head there. The toward-head cosine $\theta$ of a member is the cosine between its drift $c^B_{m,g}-c_{m,0}$ and the direction from $c_{m,0}$ to the head's generation-0 centroid, averaged over the seven members other than the head.

\paragraph{Pool reweighting.} To split a pool separation into a part due to the shares and a part due to the members, we form the share-weighted average of member centroids, normalized to unit length. With the uniform arm's members and the concentrated arm's shares this average isolates reweighting, and its distance from the uniform arm's own average is the reweighting-only column of Table~\ref{tab:rev_pool}. The share-weighted average of member centroids lies within \slot{app.recon}{0.06} of the stored pool centroid in every arm without human text, at every generation.

\FloatBarrier
\section{Full readings}
\label{app:tables}

The tables below give every reading behind the figures and the text, grouped by what they measure. All use DeBERTa-v3 except Table~\ref{tab:rev_ppl}, and Tables~\ref{tab:rev_encoder2} and~\ref{tab:encgeom} add all-mpnet-base-v2.

\subsection{Drift and pace}

Tables~\ref{tab:driftgen} to~\ref{tab:pairs} give drift and separation by generation, the uniform arm run on to generation 10 and each member's drift. Table~\ref{tab:arms} collects the generation-5 readings of every arm, Table~\ref{tab:rev_block} the spread and co-movement, Table~\ref{tab:rev_offpath} the effective generation by generation, and Table~\ref{tab:permember} the same readings member by member.

\begin{table}[t]
\caption{Drift by generation, cosine distances in units of $10^{-3}$. Member drift $D$ averages the eight members' distances to their generation-0 centroids. Pool drift is the distance of the pool centroid from the arm's generation-0 pool centroid and is left blank for the human arms, whose pools contain human text.}
\label{tab:driftgen}
\centering
\small
\begin{tabular}{llrrrrrrrrrr}
\toprule
 & & \multicolumn{5}{c}{Member drift $D$} & \multicolumn{5}{c}{Pool drift} \\
\cmidrule(lr){3-7}\cmidrule(lr){8-12}
Arm & Seed & 1 & 2 & 3 & 4 & 5 & 1 & 2 & 3 & 4 & 5 \\
\midrule
Uniform & 42 & 15.3 & 35.6 & 54.9 & 71.8 & 83.0 & 11.5 & 29.3 & 48.0 & 66.3 & 78.4 \\
Uniform & 43 & 13.9 & 35.8 & 57.4 & 72.8 & 82.5 & 11.3 & 31.0 & 49.4 & 65.7 & 76.8 \\
28\% & 42 & 13.6 & 32.8 & 51.7 & 67.6 & 80.0 & 11.3 & 28.4 & 46.3 & 62.4 & 80.2 \\
28\% & 43 & 12.4 & 32.0 & 49.5 & 63.0 & 73.9 & 10.7 & 28.2 & 41.9 & 58.3 & 71.0 \\
50\% & 42 & 11.1 & 29.2 & 46.4 & 58.2 & 66.5 & 9.3 & 26.8 & 40.0 & 50.6 & 60.1 \\
50\% & 43 & 10.9 & 28.4 & 44.5 & 54.7 & 63.4 & 9.5 & 24.8 & 37.2 & 47.6 & 52.8 \\
90\% & 42 & 8.6 & 18.6 & 24.1 & 33.7 & 38.7 & 6.3 & 13.3 & 22.1 & 22.9 & 29.7 \\
90\% & 43 & 8.4 & 16.6 & 23.0 & 32.6 & 40.8 & 4.1 & 8.9 & 19.2 & 23.7 & 35.4 \\
SmolLM2 50\% & 42 & 16.7 & 36.7 & 54.5 & 69.0 & 77.5 & 5.7 & 19.4 & 34.3 & 45.6 & 51.9 \\
SmolLM2 50\% & 43 & 17.1 & 37.2 & 54.3 & 69.1 & 75.0 & 6.6 & 18.5 & 33.4 & 45.1 & 47.0 \\
Qwen3 50\% & 42 & 18.5 & 40.9 & 67.5 & 85.7 & 93.9 & 10.9 & 37.2 & 63.2 & 79.5 & 85.2 \\
Qwen3 50\% & 43 & 17.7 & 41.6 & 68.3 & 85.0 & 94.3 & 11.7 & 35.8 & 63.4 & 79.6 & 85.8 \\
25\% human & 42 & 11.0 & 26.4 & 39.2 & 50.5 & 53.3 &  &  &  &  &  \\
50\% human & 42 & 9.4 & 20.6 & 28.4 & 35.1 & 38.9 &  &  &  &  &  \\
\bottomrule
\end{tabular}
\end{table}

\begin{table}[t]
\caption{The uniform arm with chain seed 42 run on to generation 10, cosine distances in units of $10^{-3}$. Direction against generation 5 is the cosine between a member's drift from its generation-0 centroid at the given generation and at generation 5, averaged over the eight members. Distance to Phi-2 averages the distances from the other seven members to Phi-2. Function words and the lowest member's distinct 4-gram fraction are as in Tables~\ref{tab:fw} and~\ref{tab:rev_text}.}
\label{tab:long}
\centering
\small
\setlength{\tabcolsep}{2.8pt}
\begin{tabular}{lrrrrrrrrrrr}
\toprule
 & \multicolumn{11}{c}{Generation} \\
\cmidrule(lr){2-12}
Quantity & 0 & 1 & 2 & 3 & 4 & 5 & 6 & 7 & 8 & 9 & 10 \\
\midrule
Member drift $D$ & 0.0 & 15.3 & 35.6 & 54.9 & 71.8 & 83.0 & 90.4 & 95.0 & 94.4 & 94.8 & 97.8 \\
Spread $W$ & 7.7 & 4.6 & 10.0 & 11.2 & 9.4 & 8.4 & 8.4 & 8.9 & 9.4 & 9.1 & 9.9 \\
Co-movement $\gamma$ &  & 0.87 & 0.91 & 0.94 & 0.96 & 0.97 & 0.97 & 0.97 & 0.97 & 0.98 & 0.97 \\
Direction against generation 5 &  & 0.755 & 0.907 & 0.974 & 0.996 & 1 & 0.998 & 0.996 & 0.993 & 0.990 & 0.986 \\
Distance to Phi-2 & 13.5 & 3.3 & 10.0 & 12.5 & 7.4 & 7.0 & 6.1 & 7.0 & 5.9 & 6.1 & 7.7 \\
Function words per 1{,}000 & 211 & 154 & 115 & 95 & 83 & 78 & 77 & 76 & 75 & 74 & 74 \\
Lowest distinct 4-grams &  & 0.995 & 0.998 & 0.994 & 0.989 & 0.982 & 0.978 & 0.969 & 0.960 & 0.954 & 0.957 \\
\bottomrule
\end{tabular}
\end{table}

\begin{table}[t]
\caption{Drift of each member at generation 5, cosine distance to its own generation-0 centroid, in units of $10^{-3}$. Columns are arms and their chain seeds, and h.\ marks the two human arms.}
\label{tab:members}
\centering
\footnotesize
\setlength{\tabcolsep}{1.5pt}
\begin{tabular}{lrrrrrrrrrrrrrr}
\toprule
Member & \multicolumn{2}{c}{Uniform} & \multicolumn{2}{c}{28\%} & \multicolumn{2}{c}{50\%} & \multicolumn{2}{c}{90\%} & \multicolumn{2}{c}{SmolLM2} & \multicolumn{2}{c}{Qwen3} & 25\% h. & 50\% h. \\
\cmidrule(lr){2-3}\cmidrule(lr){4-5}\cmidrule(lr){6-7}\cmidrule(lr){8-9}\cmidrule(lr){10-11}\cmidrule(lr){12-13}
Seed & 42 & 43 & 42 & 43 & 42 & 43 & 42 & 43 & 42 & 43 & 42 & 43 & 42 & 42 \\
\midrule
Qwen3-1.7B & 80.4 & 81.2 & 83.4 & 78.8 & 75.8 & 68.7 & 44.8 & 48.8 & 80.3 & 86.0 & 81.8 & 85.3 & 63.0 & 51.8 \\
Qwen2.5-1.5B & 93.0 & 93.9 & 87.9 & 82.4 & 76.7 & 77.0 & 47.2 & 45.9 & 85.7 & 83.7 & 98.4 & 97.0 & 66.4 & 52.9 \\
SmolLM2-1.7B & 35.7 & 29.9 & 32.3 & 31.4 & 29.9 & 26.7 & 16.7 & 17.7 & 31.1 & 25.8 & 33.8 & 32.3 & 15.7 & 5.8 \\
SmolLM3-3B & 71.4 & 74.0 & 67.2 & 60.2 & 46.6 & 46.9 & 20.8 & 25.0 & 60.3 & 59.3 & 97.9 & 98.5 & 37.2 & 23.6 \\
OLMo-2-1B & 124.2 & 129.1 & 120.2 & 117.5 & 108.5 & 111.5 & 72.6 & 74.1 & 119.0 & 120.0 & 132.4 & 135.6 & 93.7 & 72.5 \\
StableLM-2-1.6B & 70.7 & 67.4 & 65.5 & 51.9 & 52.0 & 47.2 & 27.4 & 28.2 & 58.2 & 55.1 & 91.6 & 91.2 & 35.7 & 22.1 \\
Phi-2 & 95.9 & 94.8 & 98.8 & 84.7 & 66.5 & 49.3 & 30.4 & 37.3 & 95.0 & 83.2 & 117.4 & 119.0 & 49.2 & 34.1 \\
Falcon3-1B & 93.1 & 89.8 & 85.0 & 84.5 & 75.8 & 79.6 & 49.6 & 49.3 & 90.8 & 87.2 & 97.5 & 95.4 & 65.8 & 48.0 \\
\midrule
Mean & 83.0 & 82.5 & 80.0 & 73.9 & 66.5 & 63.4 & 38.7 & 40.8 & 77.5 & 75.0 & 93.9 & 94.3 & 53.3 & 38.9 \\
\bottomrule
\end{tabular}
\end{table}

\begin{table}[t]
\caption{Pairs of arms by generation, cosine distances in units of $10^{-3}$. $S$, $S/D$, $\kappa$ and $\tau$ are on the member scale as in Table~\ref{tab:arms}. The pool columns give the separation between pool centroids and its ratio to the mean pool drift, for pairs without human text. The last block gives the seed floor $F$ of the uniform arm, the separation between its two runs.}
\label{tab:pairs}
\centering
\scriptsize
\setlength{\tabcolsep}{4pt}
\begin{tabular}{llrrrrrr}
\toprule
Pair & Gen. & $S$ & $S/D$ (\%) & $\kappa$ & $\tau$ & Pool $S$ & Pool $S/D$ (\%) \\
\midrule
Uniform vs 28\% & 1 & 0.26 & 1.8 & 0.990 & 0.93 & 0.32 & 2.8 \\
 & 2 & 0.28 & 0.8 & 0.996 & 1.94 & 0.43 & 1.5 \\
 & 3 & 0.43 & 0.8 & 0.996 & 2.93 & 0.47 & 1.0 \\
 & 4 & 0.71 & 1.0 & 0.996 & 3.91 & 0.84 & 1.3 \\
 & 5 & 1.07 & 1.3 & 0.994 & 4.61 & 1.08 & 1.4 \\
\addlinespace[2pt]
Uniform vs 28\% (43) & 1 & 0.28 & 2.1 & 0.989 & 0.92 & 0.39 & 3.6 \\
 & 2 & 0.38 & 1.1 & 0.994 & 1.87 & 0.65 & 2.2 \\
 & 3 & 0.96 & 1.8 & 0.994 & 2.75 & 1.45 & 3.2 \\
 & 4 & 1.33 & 2.0 & 0.993 & 3.46 & 1.66 & 2.7 \\
 & 5 & 2.32 & 3.0 & 0.986 & 4.14 & 2.24 & 3.0 \\
\addlinespace[2pt]
Uniform vs 50\% & 1 & 1.08 & 8.2 & 0.969 & 0.84 & 2.44 & 23.5 \\
 & 2 & 0.75 & 2.3 & 0.992 & 1.73 & 2.59 & 9.2 \\
 & 3 & 1.11 & 2.2 & 0.992 & 2.66 & 3.69 & 8.4 \\
 & 4 & 2.27 & 3.5 & 0.988 & 3.35 & 5.97 & 10.2 \\
 & 5 & 3.95 & 5.3 & 0.981 & 3.98 & 7.20 & 10.4 \\
\addlinespace[2pt]
Uniform vs 50\% (43) & 1 & 0.80 & 6.5 & 0.973 & 0.86 & 2.06 & 19.8 \\
 & 2 & 1.13 & 3.5 & 0.988 & 1.72 & 3.61 & 12.9 \\
 & 3 & 2.10 & 4.1 & 0.986 & 2.49 & 6.13 & 14.1 \\
 & 4 & 3.76 & 5.9 & 0.978 & 2.93 & 7.34 & 13.0 \\
 & 5 & 5.45 & 7.5 & 0.972 & 3.54 & 11.71 & 18.1 \\
\addlinespace[2pt]
Uniform vs 90\% & 1 & 3.38 & 28.3 & 0.864 & 0.67 & 11.28 & 126.8 \\
 & 2 & 6.54 & 24.2 & 0.922 & 1.18 & 20.42 & 95.8 \\
 & 3 & 12.83 & 32.5 & 0.894 & 1.47 & 28.33 & 80.8 \\
 & 4 & 17.50 & 33.2 & 0.886 & 1.91 & 46.67 & 104.6 \\
 & 5 & 24.61 & 40.4 & 0.842 & 2.06 & 61.97 & 114.7 \\
\addlinespace[2pt]
Uniform vs 90\% (43) & 1 & 3.10 & 27.8 & 0.840 & 0.65 & 13.48 & 175.2 \\
 & 2 & 8.38 & 32.0 & 0.889 & 1.08 & 27.16 & 136.1 \\
 & 3 & 15.77 & 39.2 & 0.877 & 1.39 & 32.35 & 94.2 \\
 & 4 & 19.40 & 36.8 & 0.871 & 1.85 & 43.58 & 97.5 \\
 & 5 & 22.82 & 37.0 & 0.855 & 2.34 & 44.09 & 78.6 \\
\addlinespace[2pt]
Uniform vs SmolLM2 & 1 & 0.27 & 1.7 & 0.992 & 1.06 & 0.73 & 8.5 \\
 & 2 & 0.31 & 0.9 & 0.996 & 2.11 & 1.40 & 5.8 \\
 & 3 & 1.05 & 1.9 & 0.988 & 3.22 & 2.63 & 6.4 \\
 & 4 & 1.30 & 1.8 & 0.990 & 4.10 & 3.66 & 6.5 \\
 & 5 & 1.48 & 1.8 & 0.990 & 4.65 & 4.08 & 6.3 \\
\addlinespace[2pt]
Uniform vs SmolLM2 (43) & 1 & 0.40 & 2.6 & 0.992 & 1.11 & 0.29 & 3.2 \\
 & 2 & 0.44 & 1.2 & 0.991 & 2.14 & 1.82 & 7.4 \\
 & 3 & 1.77 & 3.2 & 0.982 & 3.12 & 4.01 & 9.7 \\
 & 4 & 2.77 & 3.9 & 0.978 & 3.88 & 4.89 & 8.8 \\
 & 5 & 3.91 & 5.0 & 0.976 & 4.32 & 6.38 & 10.3 \\
\addlinespace[2pt]
Uniform vs Qwen3 & 1 & 0.45 & 2.7 & 0.988 & 1.18 & 0.58 & 5.2 \\
 & 2 & 0.62 & 1.6 & 0.993 & 2.32 & 3.24 & 9.8 \\
 & 3 & 1.89 & 3.1 & 0.990 & 3.74 & 4.59 & 8.3 \\
 & 4 & 2.22 & 2.8 & 0.991 & 4.81 & 3.88 & 5.3 \\
 & 5 & 3.54 & 4.0 & 0.983 & 4.96 & 5.56 & 6.8 \\
\addlinespace[2pt]
Uniform vs Qwen3 (43) & 1 & 0.58 & 3.7 & 0.990 & 1.23 & 0.87 & 7.6 \\
 & 2 & 1.10 & 2.9 & 0.984 & 2.37 & 3.18 & 9.5 \\
 & 3 & 1.59 & 2.5 & 0.991 & 3.64 & 4.62 & 8.2 \\
 & 4 & 2.77 & 3.5 & 0.988 & 4.83 & 4.42 & 6.1 \\
 & 5 & 2.72 & 3.1 & 0.988 & 5.00 & 4.80 & 5.9 \\
\addlinespace[2pt]
50\% vs 25\% human & 1 & 0.17 & 1.5 & 0.990 & 0.99 &  &  \\
 & 2 & 0.30 & 1.1 & 0.995 & 1.87 &  &  \\
 & 3 & 0.82 & 1.9 & 0.994 & 2.64 &  &  \\
 & 4 & 1.02 & 1.9 & 0.993 & 3.35 &  &  \\
 & 5 & 2.38 & 4.0 & 0.986 & 3.64 &  &  \\
\addlinespace[2pt]
50\% vs 50\% human & 1 & 0.42 & 4.1 & 0.974 & 0.87 &  &  \\
 & 2 & 1.49 & 6.0 & 0.990 & 1.58 &  &  \\
 & 3 & 3.75 & 10.0 & 0.982 & 2.03 &  &  \\
 & 4 & 5.31 & 11.4 & 0.974 & 2.41 &  &  \\
 & 5 & 7.73 & 14.7 & 0.961 & 2.63 &  &  \\
\addlinespace[2pt]
Uniform 42 vs 43 ($F$) & 1 & 0.23 &  &  &  & 0.07 &  \\
 & 2 & 0.28 &  &  &  & 0.03 &  \\
 & 3 & 0.31 &  &  &  & 0.13 &  \\
 & 4 & 0.53 &  &  &  & 0.29 &  \\
 & 5 & 0.79 &  &  &  & 0.39 &  \\
\bottomrule
\end{tabular}
\end{table}

\begin{table}[t]
\caption{Readings at generation 5 on the member scale, cosine distances in units of $10^{-3}$. $D$ is member drift, and head drift is that of the head, which is Phi-2 except in the SmolLM2 and Qwen3 arms. $S$ is the separation from the reference arm with the same chain seed, $S/D$ divides it by the two arms' mean drift, and chord is $\sqrt{S/D}$. $\kappa$ is the direction cosine, and $\tau$, the lag and the off-path distance place the arm on the reference arm's path. The reference is the uniform arm in the upper block and the 50\% arm with seed 42 for the human arms. Each uniform row places its run on the other uniform run's path, the seed-43 row gives the seed floor $F$ as $S$, and each row's $\tau$ is the control for arms of the other chain seed. The human arms' control is the 50\% run with seed 43 on the seed-42 path, at $\tau=4.34$. $^{\dagger}$Phi-2 as an ordinary member.}
\label{tab:arms}
\centering
\small
\setlength{\tabcolsep}{3.6pt}
\begin{tabular}{llrrrrrrrrr}
\toprule
Arm & Seed & $D$ & Head drift & $S$ & $S/D$ (\%) & Chord (\%) & $\kappa$ & $\tau$ & Lag & Off-path \\
\midrule
Uniform & 42 & 83.0 & 95.9$^{\dagger}$ &  &  &  &  & 4.81 &  & 0.70 \\
Uniform & 43 & 82.5 & 94.8$^{\dagger}$ & 0.79 & 1.0 & 10 & 0.993 & 4.72 &  & 0.67 \\
28\% & 42 & 80.0 & 98.8 & 1.07 & 1.3 & 11 & 0.994 & 4.61 & 0.11 & 0.89 \\
28\% & 43 & 73.9 & 84.7 & 2.32 & 3.0 & 17 & 0.986 & 4.14 & 0.67 & 1.51 \\
50\% & 42 & 66.5 & 66.5 & 3.95 & 5.3 & 23 & 0.981 & 3.98 & 0.75 & 1.98 \\
50\% & 43 & 63.4 & 49.3 & 5.45 & 7.5 & 27 & 0.972 & 3.54 & 1.28 & 2.47 \\
90\% & 42 & 38.7 & 30.4 & 24.61 & 40.4 & 64 & 0.842 & 2.06 & 2.66 & 7.96 \\
90\% & 43 & 40.8 & 37.3 & 22.82 & 37.0 & 61 & 0.855 & 2.34 & 2.48 & 8.22 \\
SmolLM2 50\% & 42 & 77.5 & 31.1 & 1.48 & 1.8 & 14 & 0.990 & 4.65 & 0.07 & 1.32 \\
SmolLM2 50\% & 43 & 75.0 & 25.8 & 3.91 & 5.0 & 22 & 0.976 & 4.32 & 0.50 & 3.28 \\
Qwen3 50\% & 42 & 93.9 & 81.8 & 3.54 & 4.0 & 20 & 0.983 & 4.96 & $-0.23$ & 3.53 \\
Qwen3 50\% & 43 & 94.3 & 85.3 & 2.72 & 3.1 & 18 & 0.988 & 5.00 & $-0.19$ & 2.72 \\
\midrule
\multicolumn{11}{l}{\emph{Reference: the 50\% arm, seed 42}} \\
25\% human & 42 & 53.3 & 49.2 & 2.38 & 4.0 & 20 & 0.986 & 3.64 & 0.70 & 0.39 \\
50\% human & 42 & 38.9 & 34.1 & 7.73 & 14.7 & 38 & 0.961 & 2.63 & 1.71 & 0.63 \\
\bottomrule
\end{tabular}
\end{table}

\begin{table}[t]
\caption{Spread and co-movement of every arm by generation. The spread $W$ (cosine distance in units of $10^{-3}$) averages the 28 member pairs, and $W$ without head averages the 21 pairs that leave out the head. The head is SmolLM2 or Qwen3 in the arms named after them and Phi-2 in every other arm, the uniform arm included. At generation 0 all arms of a chain seed share $W$, 7.7 with seed 42 and 7.2 with seed 43. The co-movement cosine $\gamma$ averages the eight members, and the last column gives the head's own $\gamma$.}
\label{tab:rev_block}
\centering
\small
\setlength{\tabcolsep}{3.4pt}
\begin{tabular}{llrrrrrrrrrrrr}
\toprule
 & & \multicolumn{5}{c}{Spread $W$} & \multicolumn{5}{c}{Co-movement $\gamma$} & $W$ w/o head & Head $\gamma$ \\
\cmidrule(lr){3-7}\cmidrule(lr){8-12}\cmidrule(lr){13-13}\cmidrule(lr){14-14}
Arm & Seed & 1 & 2 & 3 & 4 & 5 & 1 & 2 & 3 & 4 & 5 & 5 & 5 \\
\midrule
Uniform & 42 & 4.6 & 10.0 & 11.2 & 9.4 & 8.4 & 0.87 & 0.91 & 0.94 & 0.96 & 0.97 & 8.8 & 0.96 \\
Uniform & 43 & 5.1 & 9.1 & 10.8 & 9.4 & 8.9 & 0.90 & 0.94 & 0.95 & 0.97 & 0.97 & 9.1 & 0.94 \\
28\% & 42 & 5.3 & 9.6 & 10.7 & 9.8 & 8.4 & 0.84 & 0.91 & 0.94 & 0.95 & 0.96 & 9.2 & 0.95 \\
28\% & 43 & 5.3 & 9.7 & 12.2 & 10.1 & 9.0 & 0.86 & 0.92 & 0.93 & 0.96 & 0.96 & 9.6 & 0.95 \\
50\% & 42 & 5.6 & 9.4 & 11.8 & 12.1 & 10.6 & 0.84 & 0.90 & 0.92 & 0.94 & 0.95 & 10.1 & 0.91 \\
50\% & 43 & 5.6 & 9.8 & 12.1 & 12.3 & 13.5 & 0.86 & 0.92 & 0.93 & 0.94 & 0.95 & 10.6 & 0.91 \\
90\% & 42 & 6.6 & 9.1 & 9.3 & 12.2 & 15.3 & 0.76 & 0.88 & 0.89 & 0.90 & 0.90 & 12.6 & 0.85 \\
90\% & 43 & 7.2 & 10.2 & 8.2 & 12.4 & 12.5 & 0.77 & 0.87 & 0.91 & 0.91 & 0.91 & 12.1 & 0.89 \\
SmolLM2 50\% & 42 & 5.5 & 10.2 & 10.7 & 10.8 & 9.9 & 0.86 & 0.92 & 0.94 & 0.95 & 0.96 & 8.5 & 0.96 \\
SmolLM2 50\% & 43 & 5.6 & 10.8 & 11.9 & 10.4 & 10.3 & 0.90 & 0.92 & 0.94 & 0.96 & 0.97 & 8.2 & 0.97 \\
Qwen3 50\% & 42 & 5.0 & 9.9 & 8.7 & 8.8 & 9.7 & 0.89 & 0.92 & 0.96 & 0.97 & 0.97 & 9.3 & 0.95 \\
Qwen3 50\% & 43 & 4.9 & 11.0 & 9.1 & 9.6 & 9.6 & 0.91 & 0.92 & 0.96 & 0.98 & 0.98 & 9.0 & 0.95 \\
50\% human & 42 & 5.8 & 9.6 & 13.4 & 14.8 & 13.9 & 0.78 & 0.90 & 0.91 & 0.91 & 0.92 & 12.9 & 0.82 \\
25\% human & 42 & 5.6 & 8.5 & 11.9 & 12.7 & 12.2 & 0.83 & 0.90 & 0.93 & 0.93 & 0.95 & 11.0 & 0.86 \\
\bottomrule
\end{tabular}
\end{table}

\begin{table}[t]
\caption{Separation from the uniform arm split along and off the uniform route, by generation (cosine distances in units of $10^{-3}$). $S$ is the separation at the same generation, and off-path is the distance to the nearest point of the uniform path of the same chain seed, also given as a percentage of $S$. $\tau$ is the effective generation, the lag is the control's $\tau$ minus the arm's, and $\Delta\tau$ is the advance of $\tau$ since the previous generation. The two Phi-2 columns give the head's own $\tau$ and off-path distance. The uniform rows are the controls, which place the uniform run with seed 43 on the seed-42 path and the run with seed 42 on the seed-43 path. The lag of each arm is taken against the control on its own path, so every arm with seed 43 is compared with the seed-42 uniform run.}
\label{tab:rev_offpath}
\centering
\small
\setlength{\tabcolsep}{3.4pt}
\begin{tabular}{llrrrrrrrr}
\toprule
 & & & \multicolumn{2}{c}{Off-path} & & & & \multicolumn{2}{c}{Phi-2} \\
\cmidrule(lr){4-5}\cmidrule(lr){9-10}
Arm & Gen. & $S$ & Distance & \% of $S$ & $\tau$ & Lag & $\Delta\tau$ & $\tau$ & Off-path \\
\midrule
90\% (42) & 1 & 3.38 & 1.11 & 33 & 0.67 & 0.26 & 0.67 & 0.43 & 0.7 \\
 & 2 & 6.54 & 1.13 & 17 & 1.18 & 0.81 & 0.51 & 0.62 & 1.4 \\
 & 3 & 12.83 & 2.12 & 17 & 1.47 & 1.59 & 0.29 & 0.77 & 4.6 \\
 & 4 & 17.50 & 4.71 & 27 & 1.91 & 2.05 & 0.44 & 0.69 & 8.5 \\
 & 5 & 24.61 & 7.96 & 32 & 2.06 & 2.66 & 0.15 & 0.65 & 16.9 \\
\addlinespace[2pt]
90\% (43) & 1 & 3.10 & 1.08 & 35 & 0.65 & 0.43 & 0.65 & 0.39 & 0.7 \\
 & 2 & 8.38 & 1.42 & 17 & 1.08 & 0.95 & 0.43 & 0.57 & 2.1 \\
 & 3 & 15.77 & 2.64 & 17 & 1.39 & 1.56 & 0.31 & 0.77 & 5.6 \\
 & 4 & 19.40 & 5.35 & 28 & 1.85 & 2.18 & 0.46 & 0.73 & 9.1 \\
 & 5 & 22.82 & 8.22 & 36 & 2.34 & 2.48 & 0.49 & 2.48 & 12.4 \\
\addlinespace[2pt]
50\% (42) & 1 & 1.08 & 0.32 & 29 & 0.84 & 0.09 & 0.84 & 0.65 & 0.2 \\
 & 2 & 0.75 & 0.29 & 38 & 1.73 & 0.26 & 0.89 & 1.32 & 0.4 \\
 & 3 & 1.11 & 0.57 & 51 & 2.66 & 0.40 & 0.94 & 2.58 & 0.7 \\
 & 4 & 2.27 & 0.80 & 35 & 3.35 & 0.61 & 0.69 & 3.06 & 0.9 \\
 & 5 & 3.95 & 1.98 & 50 & 3.98 & 0.75 & 0.63 & 3.50 & 2.0 \\
\addlinespace[2pt]
50\% (43) & 1 & 0.80 & 0.35 & 44 & 0.86 & 0.22 & 0.86 & 0.68 & 0.3 \\
 & 2 & 1.13 & 0.50 & 45 & 1.72 & 0.32 & 0.86 & 1.37 & 1.3 \\
 & 3 & 2.10 & 0.76 & 36 & 2.49 & 0.47 & 0.77 & 2.31 & 3.0 \\
 & 4 & 3.76 & 1.30 & 35 & 2.93 & 1.10 & 0.44 & 2.97 & 2.9 \\
 & 5 & 5.45 & 2.47 & 45 & 3.54 & 1.28 & 0.61 & 3.00 & 10.6 \\
\addlinespace[2pt]
Uniform (43) & 1 & 0.23 & 0.12 & 53 & 0.93 &  & 0.92 & 0.89 & 0.3 \\
 & 2 & 0.28 & 0.23 & 83 & 1.99 &  & 1.06 & 2.08 & 0.2 \\
 & 3 & 0.31 & 0.26 & 84 & 3.06 &  & 1.07 & 3.16 & 0.7 \\
 & 4 & 0.53 & 0.43 & 81 & 3.96 &  & 0.90 & 3.87 & 1.0 \\
 & 5 & 0.79 & 0.67 & 85 & 4.72 &  & 0.76 & 4.32 & 1.1 \\
\addlinespace[2pt]
Uniform (42) & 1 & 0.23 & 0.14 & 61 & 1.08 &  & 1.04 & 1.27 & 0.2 \\
 & 2 & 0.28 & 0.23 & 82 & 2.04 &  & 0.96 & 1.94 & 0.2 \\
 & 3 & 0.31 & 0.27 & 88 & 2.96 &  & 0.92 & 2.87 & 0.8 \\
 & 4 & 0.53 & 0.47 & 90 & 4.03 &  & 1.08 & 4.51 & 1.1 \\
 & 5 & 0.79 & 0.70 & 88 & 4.81 &  & 0.78 & 5.00 & 1.4 \\
\bottomrule
\end{tabular}
\end{table}

\begin{table}[htbp]
\caption{Per-member readings at generation 5 for the 50\% and 90\% Phi-2 arms, each member placed on its own path in the uniform arm with the same chain seed (cosine distances in units of $10^{-3}$). $S$ is the member's separation from itself in the uniform arm, Delay and Off are its delay and off-route parts beyond the same member's parts in the control, and the lag is the member's effective generation in the control minus its effective generation in the arm, with the control as in Table~\ref{tab:split}. The row below each block counts the members whose delay beyond the control is at least their off-route part beyond it. At 90\% the cross term is negative for 4 and 3 of the 8 members in the two runs, and no member's nearest point lies at the end of its path.}
\label{tab:permember}
\centering
\small
\setlength{\tabcolsep}{4pt}
\begin{tabular}{lrrrrrrrr}
\toprule
 & \multicolumn{4}{c}{50\% Phi-2} & \multicolumn{4}{c}{90\% Phi-2} \\
\cmidrule(lr){2-5}\cmidrule(lr){6-9}
Member & $S$ & Delay & Off & Lag & $S$ & Delay & Off & Lag \\
\midrule
\multicolumn{9}{l}{\emph{Chain seed 42}} \\
Qwen3-1.7B & 4.9 & 0.9 & 1.3 & 1.00 & 18.3 & 7.5 & 5.9 & 2.46 \\
Qwen2.5-1.5B & 4.1 & 0.1 & 3.5 & 0.11 & 19.1 & 13.3 & 6.9 & 2.36 \\
SmolLM2-1.7B & 1.1 & 0.6 & $-$0.3 & 0.71 & 10.0 & 7.3 & 1.2 & 2.29 \\
SmolLM3-3B & 5.1 & 3.3 & 0.6 & 1.11 & 39.7 & 34.4 & 6.5 & 2.94 \\
OLMo-2-1B & 2.2 & $-$0.4 & 1.5 & $-$0.68 & 16.5 & 13.7 & 9.0 & 2.27 \\
StableLM-2-1.6B & 4.1 & 2.5 & 1.2 & 0.91 & 30.6 & 16.8 & 9.7 & 2.32 \\
Phi-2 & 6.3 & 4.0 & 0.9 & 0.82 & 46.9 & 49.6 & 15.8 & 3.67 \\
Falcon3-1B & 3.8 & 1.7 & 1.7 & 2.00 & 15.8 & 10.3 & 3.3 & 3.00 \\
Delay $\ge$ off & \multicolumn{4}{c}{4 of 8} & \multicolumn{4}{c}{8 of 8} \\
\midrule
\multicolumn{9}{l}{\emph{Chain seed 43}} \\
Qwen3-1.7B & 4.6 & 1.2 & 0.6 & 0.58 & 22.5 & 3.9 & 9.6 & 1.19 \\
Qwen2.5-1.5B & 3.3 & 1.8 & 1.2 & 1.21 & 22.5 & 12.9 & 7.5 & 2.56 \\
SmolLM2-1.7B & 1.6 & 1.1 & $-$0.1 & 2.00 & 7.8 & 2.4 & 2.2 & 2.38 \\
SmolLM3-3B & 6.2 & 4.1 & 0.7 & 1.25 & 36.8 & 27.3 & 9.1 & 2.72 \\
OLMo-2-1B & 1.9 & 0.4 & 0.4 & 0.75 & 21.2 & 15.5 & 9.4 & 3.00 \\
StableLM-2-1.6B & 5.8 & 2.6 & 1.7 & 1.06 & 28.7 & 21.5 & 8.7 & 2.77 \\
Phi-2 & 18.6 & 6.3 & 9.2 & 2.00 & 29.1 & 12.3 & 11.0 & 2.52 \\
Falcon3-1B & 1.6 & 0.8 & 0.5 & 1.36 & 14.0 & 10.6 & 2.7 & 2.65 \\
Delay $\ge$ off & \multicolumn{4}{c}{6 of 8} & \multicolumn{4}{c}{7 of 8} \\
\bottomrule
\end{tabular}
\end{table}

\FloatBarrier
\subsection{Distance to the head and direction}

Table~\ref{tab:rev_headdist} gives the distance from the other members to the head by generation, Table~\ref{tab:rev_toward} the cosines between the members' drift or departure and the direction to the head, Table~\ref{tab:towardnull} the departure cosine toward every member, and Table~\ref{tab:rev_kappa} the direction cosine member by member.

\begin{table}[t]
\caption{Distance from the other seven members to the head at the same generation (cosine distance in units of $10^{-3}$, averaged over the seven). The head is Phi-2 except in the SmolLM2 and Qwen3 arms, and the uniform arm is given for all three references. Generation 0 is shared by all arms of a chain seed. Matched gives the same distance at generation 5 after placing every member, head included, at its nearest point on the uniform arm's path of the same chain seed. For uniform seed 43 it places the seed-43 members on the seed-42 uniform path, so its gap between observed and matched values is the gap expected without concentration. The last two columns give, at generation 5, how much more the distance fell than in the uniform arm with the same seed and reference, and that excess divided by the difference between the two uniform seeds in the same quantity, which is Phi-2 for every row except the SmolLM2 and Qwen3 arms. A negative excess means the members ended farther from the head than in the uniform arm. Matched is left blank for the seed-42 uniform run, whose path it is, and for the two human arms.}
\label{tab:rev_headdist}
\centering
\small
\setlength{\tabcolsep}{3.6pt}
\begin{tabular}{lllrrrrrrrrr}
\toprule
 & & & \multicolumn{6}{c}{Generation} & & \multicolumn{2}{c}{Excess fall at 5} \\
\cmidrule(lr){4-9}\cmidrule(lr){11-12}
Arm & Seed & Head & 0 & 1 & 2 & 3 & 4 & 5 & Matched & Distance & / floor \\
\midrule
Uniform & 42 & Phi-2 & 13.5 & 3.3 & 10.0 & 12.5 & 7.4 & 7.0 &  &  &  \\
Uniform & 42 & SmolLM2 & 5.7 & 5.3 & 12.5 & 12.2 & 13.0 & 12.7 &  &  &  \\
Uniform & 42 & Qwen3 & 10.9 & 4.0 & 10.9 & 13.4 & 10.1 & 8.4 &  &  &  \\
Uniform & 43 & Phi-2 & 12.1 & 4.2 & 9.4 & 10.5 & 7.4 & 8.2 & 8.3 &  &  \\
Uniform & 43 & SmolLM2 & 4.9 & 4.5 & 10.1 & 12.7 & 14.9 & 16.0 & 13.3 &  &  \\
Uniform & 43 & Qwen3 & 9.5 & 4.8 & 9.8 & 11.8 & 10.0 & 8.4 & 9.7 &  &  \\
28\% & 42 & Phi-2 & 13.5 & 3.9 & 8.9 & 9.3 & 9.8 & 5.9 & 6.3 & 1.1 & 0.4 \\
28\% & 43 & Phi-2 & 12.1 & 4.6 & 9.1 & 14.7 & 9.2 & 7.3 & 7.3 & 0.9 & 0.4 \\
50\% & 42 & Phi-2 & 13.5 & 6.2 & 9.8 & 13.5 & 14.7 & 12.0 & 12.2 & $-$5.0 & $-$1.9 \\
50\% & 43 & Phi-2 & 12.1 & 6.5 & 11.2 & 15.7 & 13.3 & 22.4 & 14.2 & $-$14.1 & $-$5.4 \\
90\% & 42 & Phi-2 & 13.5 & 8.5 & 12.6 & 11.0 & 19.2 & 23.3 & 30.0 & $-$16.3 & $-$6.2 \\
90\% & 43 & Phi-2 & 12.1 & 10.2 & 13.5 & 10.7 & 17.5 & 13.6 & 8.5 & $-$5.3 & $-$2.0 \\
SmolLM2 50\% & 42 & SmolLM2 & 5.7 & 7.7 & 11.8 & 13.6 & 16.4 & 14.2 & 15.7 & $-$1.5 & $-$0.4 \\
SmolLM2 50\% & 43 & SmolLM2 & 4.9 & 7.3 & 13.9 & 13.1 & 15.2 & 16.7 & 20.4 & $-$0.7 & $-$0.2 \\
Qwen3 50\% & 42 & Qwen3 & 10.9 & 3.4 & 11.9 & 11.7 & 9.2 & 10.8 & 8.5 & $-$2.3 & $-$1.7 \\
Qwen3 50\% & 43 & Qwen3 & 9.5 & 3.8 & 11.1 & 10.2 & 9.1 & 11.4 & 8.4 & $-$3.0 & $-$2.1 \\
50\% human & 42 & Phi-2 & 13.5 & 5.1 & 14.3 & 19.8 & 22.5 & 16.9 &  & $-$9.9 & $-$3.8 \\
25\% human & 42 & Phi-2 & 13.5 & 5.6 & 8.2 & 14.0 & 15.9 & 15.8 &  & $-$8.7 & $-$3.3 \\
\midrule
\multicolumn{12}{l}{Floor, difference in the change since generation 0 between uniform seeds 42 and 43, by generation} \\
\multicolumn{12}{l}{Phi-2: 1: 2.30, 2: 0.78, 3: 0.60, 4: 1.41, 5: 2.61} \\
\multicolumn{12}{l}{SmolLM2: 1: 0.09, 2: 1.68, 3: 1.26, 4: 2.66, 5: 4.05} \\
\multicolumn{12}{l}{Qwen3: 1: 2.16, 2: 0.28, 3: 0.28, 4: 1.28, 5: 1.40} \\
\bottomrule
\end{tabular}
\end{table}

\begin{table}[t]
\caption{Toward-head cosines, averaged over the seven members other than the head. The first five columns give the toward-head cosine $\theta$ by generation, the cosine between a member's drift from its generation-0 centroid and the direction from that centroid to the head's generation-0 centroid. Pool uses instead the head's centroid one generation earlier, whose text the members were fine-tuned on. The last two columns take at generation 5 the departure of each member from its nearest point on the uniform route of the same chain seed and give its cosine with the direction from that point to the head's current centroid (now) and with the direction from the member's generation-0 centroid to the head's (start). For the uniform arm the departure is measured from the uniform route of the other chain seed, and for the 50\% human arm from the route of the 50\% arm, and the 25\% human arm has no departure columns. The head is Phi-2, except in the SmolLM2 and Qwen3 arms and in the uniform rows marked with another member. Under all-mpnet-base-v2, $\theta$ toward Phi-2 at generation 5 is 0.00 in the uniform arm with either seed and 0.20 in the 90\% arm, and the departure of the 90\% arm points toward Phi-2's current centroid with a cosine of 0.59, against 0.18 for the uniform control.}
\label{tab:rev_toward}
\centering
\small
\setlength{\tabcolsep}{4pt}
\begin{tabular}{lllrrrrrrrr}
\toprule
 & & & \multicolumn{5}{c}{$\theta$ by generation} & & \multicolumn{2}{c}{Departure at 5} \\
\cmidrule(lr){4-8}\cmidrule(lr){10-11}
Arm & Seed & Head & 1 & 2 & 3 & 4 & 5 & Pool 5 & Now & Start \\
\midrule
Uniform & 42 & Phi-2 & $-$0.51 & $-$0.53 & $-$0.50 & $-$0.47 & $-$0.46 & 0.96 & 0.24 & 0.10 \\
Uniform & 42 & SmolLM2 & 0.60 & 0.47 & 0.39 & 0.32 & 0.28 & 0.96 & $-$0.02 & $-$0.27 \\
Uniform & 42 & Qwen3 & 0.76 & 0.74 & 0.72 & 0.69 & 0.67 & 0.98 & 0.53 & $-$0.15 \\
Uniform & 43 & Phi-2 & $-$0.53 & $-$0.57 & $-$0.52 & $-$0.50 & $-$0.49 & 0.96 & 0.03 & $-$0.18 \\
Uniform & 43 & SmolLM2 & 0.52 & 0.41 & 0.35 & 0.30 & 0.28 & 0.96 & $-$0.01 & 0.21 \\
Uniform & 43 & Qwen3 & 0.71 & 0.68 & 0.66 & 0.64 & 0.63 & 0.98 & 0.24 & 0.24 \\
28\% & 42 & Phi-2 & $-$0.44 & $-$0.51 & $-$0.48 & $-$0.45 & $-$0.41 & 0.94 & 0.36 & 0.37 \\
28\% & 43 & Phi-2 & $-$0.46 & $-$0.52 & $-$0.49 & $-$0.47 & $-$0.43 & 0.94 & 0.54 & 0.33 \\
50\% & 42 & Phi-2 & $-$0.38 & $-$0.49 & $-$0.47 & $-$0.44 & $-$0.40 & 0.89 & 0.61 & 0.45 \\
50\% & 43 & Phi-2 & $-$0.40 & $-$0.51 & $-$0.49 & $-$0.45 & $-$0.44 & 0.89 & 0.59 & 0.35 \\
90\% & 42 & Phi-2 & $-$0.15 & $-$0.38 & $-$0.38 & $-$0.34 & $-$0.29 & 0.71 & 0.72 & 0.48 \\
90\% & 43 & Phi-2 & $-$0.13 & $-$0.32 & $-$0.36 & $-$0.34 & $-$0.30 & 0.71 & 0.73 & 0.50 \\
SmolLM2 50\% & 42 & SmolLM2 & 0.60 & 0.45 & 0.34 & 0.28 & 0.24 & 0.94 & 0.38 & $-$0.37 \\
SmolLM2 50\% & 43 & SmolLM2 & 0.52 & 0.41 & 0.32 & 0.26 & 0.24 & 0.95 & 0.48 & $-$0.27 \\
Qwen3 50\% & 42 & Qwen3 & 0.81 & 0.74 & 0.71 & 0.67 & 0.67 & 0.98 & 0.78 & 0.27 \\
Qwen3 50\% & 43 & Qwen3 & 0.73 & 0.68 & 0.65 & 0.61 & 0.62 & 0.98 & 0.76 & 0.22 \\
50\% human & 42 & Phi-2 & $-$0.32 & $-$0.50 & $-$0.50 & $-$0.50 & $-$0.47 & 0.69 & 0.13 & 0.18 \\
25\% human & 42 & Phi-2 & $-$0.39 & $-$0.49 & $-$0.50 & $-$0.47 & $-$0.47 & 0.87 &  &  \\
\bottomrule
\end{tabular}
\end{table}

\begin{table}[htbp]
\caption{Direction of the departure from the uniform route at generation 5, toward the head and toward every other member. For each member $j$ we take the cosine between every other member's departure from its nearest route point and the vector from that point to member $j$'s current centroid, and average it over the seven members other than $j$. Head gives this cosine for $j$ the head (Phi-2, or the member named in the row), Others gives its smallest and largest value over the seven members other than the head, and Rank places the head among all eight (1 is the largest). The uniform rows are the two controls, read with Phi-2 as the head. Each arm is placed on the uniform path of its own chain seed.}
\label{tab:towardnull}
\centering
\small
\setlength{\tabcolsep}{4pt}
\begin{tabular}{lrrrrrrrr}
\toprule
 & \multicolumn{4}{c}{DeBERTa-v3} & \multicolumn{4}{c}{all-mpnet-base-v2} \\
\cmidrule(lr){2-5}\cmidrule(lr){6-9}
 & & \multicolumn{2}{c}{Others} & & & \multicolumn{2}{c}{Others} & \\
\cmidrule(lr){3-4}\cmidrule(lr){7-8}
Arm, seed & Head & Min & Max & Rank & Head & Min & Max & Rank \\
\midrule
Uniform, 43 (control) & 0.03 & $-$0.23 & 0.61 & 5 & 0.18 & 0.15 & 0.32 & 5 \\
Uniform, 42 (control) & 0.24 & $-$0.28 & 0.53 & 5 & 0.12 & $-$0.01 & 0.48 & 7 \\
28\% Phi-2, 42 & 0.36 & $-$0.27 & 0.66 & 3 & 0.33 & 0.01 & 0.41 & 3 \\
28\% Phi-2, 43 & 0.54 & $-$0.07 & 0.74 & 3 & 0.40 & 0.16 & 0.45 & 3 \\
50\% Phi-2, 42 & 0.61 & $-$0.08 & 0.63 & 2 & 0.39 & 0.15 & 0.44 & 2 \\
50\% Phi-2, 43 & 0.59 & $-$0.04 & 0.58 & 1 & 0.50 & 0.23 & 0.53 & 3 \\
90\% Phi-2, 42 & 0.72 & 0.43 & 0.66 & 1 & 0.59 & 0.38 & 0.61 & 2 \\
90\% Phi-2, 43 & 0.73 & 0.37 & 0.70 & 1 & 0.63 & 0.40 & 0.66 & 3 \\
50\% SmolLM2, 42 & 0.38 & $-$0.30 & 0.66 & 4 & 0.27 & 0.25 & 0.39 & 7 \\
50\% SmolLM2, 43 & 0.48 & $-$0.10 & 0.79 & 6 & 0.21 & 0.27 & 0.45 & 8 \\
50\% Qwen3, 42 & 0.78 & $-$0.29 & 0.78 & 1 & 0.74 & 0.00 & 0.72 & 1 \\
50\% Qwen3, 43 & 0.76 & $-$0.34 & 0.79 & 2 & 0.76 & $-$0.03 & 0.73 & 1 \\
25\% human, 42 & 0.42 & 0.03 & 0.46 & 2 & 0.41 & 0.35 & 0.50 & 6 \\
50\% human, 42 & 0.39 & 0.19 & 0.50 & 4 & 0.48 & 0.29 & 0.53 & 4 \\
\bottomrule
\end{tabular}
\end{table}

\begin{table}[t]
\caption{Direction cosine of each member at generation 5, between its drift in an arm and its drift in the reference arm with the same chain seed. Columns are arms and their chain seeds. The reference is the uniform arm, except for the 50\% human arm, whose reference is the 50\% arm. The control column compares the two chain seeds of the uniform arm, each member measured from its own generation-0 centroid. The Mean row is $\kappa$.}
\label{tab:rev_kappa}
\centering
\footnotesize
\setlength{\tabcolsep}{2pt}
\begin{tabular}{lrrrrrrrrrrrr}
\toprule
Member & \multicolumn{2}{c}{28\%} & \multicolumn{2}{c}{50\%} & \multicolumn{2}{c}{90\%} & \multicolumn{2}{c}{SmolLM2} & \multicolumn{2}{c}{Qwen3} & 50\% human & Control \\
\cmidrule(lr){2-3}\cmidrule(lr){4-5}\cmidrule(lr){6-7}\cmidrule(lr){8-9}\cmidrule(lr){10-11}
Seed & 42 & 43 & 42 & 43 & 42 & 43 & 42 & 43 & 42 & 43 & 42 &  \\
\midrule
Qwen3-1.7B & 0.988 & 0.973 & 0.969 & 0.973 & 0.890 & 0.854 & 0.991 & 0.956 & 0.935 & 0.965 & 0.980 & 0.983 \\
Qwen2.5-1.5B & 0.990 & 0.986 & 0.980 & 0.985 & 0.914 & 0.894 & 0.994 & 0.974 & 0.985 & 0.987 & 0.970 & 0.997 \\
SmolLM2-1.7B & 0.995 & 0.975 & 0.987 & 0.974 & 0.869 & 0.864 & 0.989 & 0.975 & 0.994 & 0.997 & 0.946 & 0.992 \\
SmolLM3-3B & 0.993 & 0.988 & 0.979 & 0.974 & 0.681 & 0.723 & 0.973 & 0.971 & 0.988 & 0.988 & 0.952 & 0.993 \\
OLMo-2-1B & 0.998 & 0.992 & 0.993 & 0.995 & 0.950 & 0.931 & 0.998 & 0.991 & 0.988 & 0.996 & 0.990 & 0.993 \\
StableLM-2-1.6B & 0.993 & 0.990 & 0.978 & 0.964 & 0.767 & 0.767 & 0.984 & 0.973 & 0.985 & 0.989 & 0.953 & 0.999 \\
Phi-2 & 0.999 & 0.991 & 0.978 & 0.918 & 0.735 & 0.866 & 0.999 & 0.975 & 0.994 & 0.986 & 0.915 & 0.993 \\
Falcon3-1B & 0.993 & 0.995 & 0.983 & 0.993 & 0.934 & 0.940 & 0.994 & 0.992 & 0.996 & 0.998 & 0.987 & 0.999 \\
\midrule
Mean & 0.994 & 0.986 & 0.981 & 0.972 & 0.842 & 0.855 & 0.990 & 0.976 & 0.983 & 0.988 & 0.961 & 0.993 \\
\bottomrule
\end{tabular}
\end{table}

\FloatBarrier
\subsection{Pool scale and the second encoder}

Table~\ref{tab:rev_pool} gives the pool scale, and Tables~\ref{tab:rev_encoder2} and~\ref{tab:encgeom} compare direction, pace and geometry under the two encoders.

\begin{table}[t]
\caption{Pool scale (cosine distances in units of $10^{-3}$). Pool distance at generation 0 is the distance between an arm's generation-0 pool centroid and the uniform arm's, the abscissa of Figure~\ref{fig:pull}. Pool $S$ is the same distance at generation 5, and pool $S/D$ divides it by the mean pool drift of the two arms, with its square root, the chord ratio, beside it. The member columns give $S/D$ and the chord ratio on the member scale, as in Table~\ref{tab:arms}. Reweighting only is the distance at generation 5 between the uniform arm's member centroids averaged with the uniform arm's shares and the same centroids averaged with the arm's shares. Check is the largest distance, over generations 0 to 5, between an arm's share-weighted average of member centroids and its stored pool centroid. The uniform rows give the check only. The human rows compare the two human arms, whose pools include the human text, with the uniform arm with seed 42, on both scales, and leave the last two columns blank because their pools are not averages of member centroids.}
\label{tab:rev_pool}
\centering
\small
\setlength{\tabcolsep}{3.1pt}
\begin{tabular}{llrrrrrrrr}
\toprule
 & & \multicolumn{4}{c}{Pool} & \multicolumn{2}{c}{Members} & Reweighting & \\
\cmidrule(lr){3-6}\cmidrule(lr){7-8}
Arm & Seed & Gen.\ 0 & $S$ & $S/D$ (\%) & Chord (\%) & $S/D$ (\%) & Chord (\%) & only & Check \\
\midrule
Uniform & 42 & & & & & & & & 0.02 \\
Uniform & 43 & & & & & & & & 0.02 \\
28\% & 42 & 0.37 & 1.08 & 1.4 & 12 & 1.3 & 11 & 0.08 & 0.03 \\
28\% & 43 & 0.24 & 2.24 & 3.0 & 17 & 3.0 & 17 & 0.11 & 0.02 \\
50\% & 42 & 1.70 & 7.20 & 10.4 & 32 & 5.3 & 23 & 0.46 & 0.02 \\
50\% & 43 & 1.28 & 11.71 & 18.1 & 43 & 7.5 & 27 & 0.62 & 0.02 \\
90\% & 42 & 6.66 & 61.97 & 114.7 & 107 & 40.4 & 64 & 1.95 & 0.06 \\
90\% & 43 & 5.64 & 44.09 & 78.6 & 89 & 37.0 & 61 & 2.63 & 0.07 \\
SmolLM2 50\% & 42 & 0.27 & 4.08 & 6.3 & 25 & 1.8 & 14 & 1.37 & 0.03 \\
SmolLM2 50\% & 43 & 0.34 & 6.38 & 10.3 & 32 & 5.0 & 22 & 1.87 & 0.03 \\
Qwen3 50\% & 42 & 0.77 & 5.56 & 6.8 & 26 & 4.0 & 20 & 0.69 & 0.05 \\
Qwen3 50\% & 43 & 0.95 & 4.80 & 5.9 & 24 & 3.1 & 18 & 0.65 & 0.02 \\
25\% human & 42 & 9.11 & 35.53 & 66.9 & 82 & 9.0 & 30 & & \\
50\% human & 42 & 21.70 & 93.07 & 216.0 & 147 & 22.2 & 47 & & \\
\bottomrule
\end{tabular}
\end{table}

\begin{table}[t]
\caption{Direction and pace at generation 5 under the paper's encoder (DeBERTa-v3) and a second, purpose-built sentence encoder (all-mpnet-base-v2), computed from the same 800 texts per member. $\kappa$ is the direction cosine against the reference arm with the same chain seed, $\tau$ the effective generation on the reference arm's path, and the lag the difference between $\tau$ of a second run of the reference arm with the other chain seed placed on the same path and $\tau$ of the arm. $S/F$ divides the separation from the reference arm by the reference arm's seed floor. The reference is the uniform arm, except in the row placed on the 50\% path, where it is the 50\% arm. The uniform seed floor is 0.79 under DeBERTa and 6.21 under mpnet, the 50\% arm's floor is 1.27 and 8.53, and the uniform arm's member drift at generation 5 is 83.0 and 290.5 (cosine distances in units of $10^{-3}$). The control row places the uniform run with seed 43 on the seed-42 uniform path.}
\label{tab:rev_encoder2}
\centering
\small
\setlength{\tabcolsep}{4pt}
\begin{tabular}{lrrrrrrrr}
\toprule
 & \multicolumn{4}{c}{DeBERTa-v3} & \multicolumn{4}{c}{all-mpnet-base-v2} \\
\cmidrule(lr){2-5}\cmidrule(lr){6-9}
Arm & $\kappa$ & $\tau$ & Lag & $S/F$ & $\kappa$ & $\tau$ & Lag & $S/F$ \\
\midrule
28\% Phi-2 & 0.994 & 4.61 & 0.11 & 1.4 & 0.991 & 4.67 & $-$0.08 & 0.8 \\
28\% Phi-2 (s43) & 0.986 & 4.14 & 0.67 & 2.9 & 0.987 & 4.50 & 0.12 & 1.1 \\
50\% Phi-2 & 0.981 & 3.98 & 0.75 & 5.0 & 0.978 & 3.69 & 0.89 & 2.2 \\
50\% Phi-2 (s43) & 0.972 & 3.54 & 1.28 & 6.9 & 0.971 & 3.74 & 0.88 & 2.6 \\
90\% Phi-2 & 0.842 & 2.06 & 2.66 & 31.0 & 0.775 & 1.62 & 2.97 & 16.6 \\
90\% Phi-2 (s43) & 0.855 & 2.34 & 2.48 & 28.8 & 0.793 & 1.79 & 2.82 & 15.0 \\
50\% SmolLM2 & 0.990 & 4.65 & 0.07 & 1.9 & 0.991 & 4.76 & $-$0.17 & 0.8 \\
50\% SmolLM2 (s43) & 0.976 & 4.32 & 0.50 & 4.9 & 0.988 & 4.63 & $-$0.01 & 1.0 \\
50\% Qwen3 & 0.983 & 4.96 & $-$0.23 & 4.5 & 0.985 & 4.90 & $-$0.31 & 1.5 \\
50\% Qwen3 (s43) & 0.988 & 5.00 & $-$0.19 & 3.4 & 0.985 & 4.91 & $-$0.29 & 1.6 \\
50\% human, on the 50\% path & 0.961 & 2.63 & 1.71 & 6.1 & 0.928 & 2.63 & 1.94 & 3.9 \\
50\% human, on the uniform path & 0.957 & 2.30 & 2.43 & 17.1 & 0.904 & 2.20 & 2.39 & 9.2 \\
\midrule
Uniform (s43), control & 0.993 & 4.72 &  & 1.0 & 0.959 & 4.59 &  & 1.0 \\
\bottomrule
\end{tabular}
\end{table}

\begin{table}[h]
\caption{Geometry at generation 5 under DeBERTa-v3 and all-mpnet-base-v2, computed from the same 800 texts per member (cosine distances in units of $10^{-3}$). $W$ is the spread among the eight members, $\gamma$ the co-movement cosine and Head the mean distance from the seven other members to the head, which is Phi-2 except in the SmolLM2 and Qwen3 rows. The generation-0 rows give the shared starting point of each chain seed. Delay and off-route are the parts of the separation from the uniform arm defined in Table~\ref{tab:split}, Off/ctrl divides the off-route part by that of the control on the same path, and Toward is the departure cosine toward the head of Table~\ref{tab:split}. The control row places the uniform run with seed 43 on the seed-42 path. $^{*}$Placed on the path of the 50\% Phi-2 arm with seed 42.}
\label{tab:encgeom}
\centering
\small
\setlength{\tabcolsep}{4pt}
\begin{tabular}{lrrrrrrrr}
\toprule
Arm & Seed & $W$ & $\gamma$ & Head & Delay & Off & Off/ctrl & Toward \\
\midrule
\multicolumn{9}{l}{\emph{DeBERTa-v3}} \\
Generation 0 & 42 & 7.7 &  & 13.5 &  &  &  &  \\
Generation 0 & 43 & 7.2 &  & 12.1 &  &  &  &  \\
Uniform & 42 & 8.4 & 0.97 & 7.0 &  &  &  &  \\
Uniform (control) & 43 & 8.9 & 0.97 & 8.2 & 0.1 & 0.7 & 1.00 & 0.03 \\
28\% Phi-2 & 42 & 8.4 & 0.96 & 5.9 & 0.2 & 0.9 & 1.32 & 0.36 \\
28\% Phi-2 & 43 & 9.0 & 0.96 & 7.3 & 0.7 & 1.5 & 2.18 & 0.54 \\
50\% Phi-2 & 42 & 10.6 & 0.95 & 12.0 & 1.7 & 2.0 & 2.94 & 0.61 \\
50\% Phi-2 & 43 & 13.5 & 0.95 & 22.4 & 2.4 & 2.5 & 3.54 & 0.59 \\
90\% Phi-2 & 42 & 15.3 & 0.90 & 23.3 & 19.2 & 8.0 & 11.83 & 0.72 \\
90\% Phi-2 & 43 & 12.5 & 0.91 & 13.6 & 13.4 & 8.2 & 11.81 & 0.73 \\
50\% SmolLM2 & 42 & 9.9 & 0.96 & 14.2 & 0.2 & 1.3 & 1.96 & 0.38 \\
50\% SmolLM2 & 43 & 10.3 & 0.97 & 16.7 & 0.4 & 3.3 & 4.72 & 0.48 \\
50\% Qwen3 & 42 & 9.7 & 0.97 & 10.8 & 0.0 & 3.5 & 5.24 & 0.78 \\
50\% Qwen3 & 43 & 9.6 & 0.98 & 11.4 & 0.0 & 2.7 & 3.91 & 0.76 \\
50\% human$^{*}$ & 42 & 13.9 & 0.92 & 16.9 & 8.1 & 0.6 & 1.14 & 0.13 \\
\midrule
\multicolumn{9}{l}{\emph{all-mpnet-base-v2}} \\
Generation 0 & 42 & 100.6 &  & 107.8 &  &  &  &  \\
Generation 0 & 43 & 99.2 &  & 106.4 &  &  &  &  \\
Uniform & 42 & 21.2 & 0.89 & 22.5 &  &  &  &  \\
Uniform (control) & 43 & 22.6 & 0.89 & 24.7 & 1.1 & 5.1 & 1.00 & 0.18 \\
28\% Phi-2 & 42 & 20.2 & 0.89 & 19.0 & 0.8 & 4.1 & 0.80 & 0.33 \\
28\% Phi-2 & 43 & 22.8 & 0.88 & 23.1 & 2.0 & 4.6 & 0.88 & 0.40 \\
50\% Phi-2 & 42 & 29.3 & 0.87 & 40.6 & 10.0 & 7.5 & 1.46 & 0.39 \\
50\% Phi-2 & 43 & 32.2 & 0.86 & 45.1 & 9.1 & 9.5 & 1.81 & 0.50 \\
90\% Phi-2 & 42 & 68.1 & 0.71 & 98.1 & 88.2 & 32.6 & 6.35 & 0.59 \\
90\% Phi-2 & 43 & 56.0 & 0.73 & 70.5 & 72.1 & 31.7 & 6.04 & 0.63 \\
50\% SmolLM2 & 42 & 20.3 & 0.89 & 30.3 & 0.4 & 4.8 & 0.94 & 0.27 \\
50\% SmolLM2 & 43 & 22.7 & 0.89 & 36.9 & 1.0 & 5.4 & 1.02 & 0.21 \\
50\% Qwen3 & 42 & 19.9 & 0.90 & 20.2 & 0.1 & 9.5 & 1.85 & 0.74 \\
50\% Qwen3 & 43 & 21.6 & 0.90 & 21.7 & 0.1 & 9.7 & 1.86 & 0.76 \\
50\% human$^{*}$ & 42 & 48.9 & 0.80 & 80.4 & 28.6 & 12.8 & 1.79 & 0.30 \\
\bottomrule
\end{tabular}
\end{table}

\FloatBarrier
\subsection{Text}

Table~\ref{tab:fw} gives the function-word rate by generation, Table~\ref{tab:rev_ppl} perplexity under GPT-2-large, Table~\ref{tab:rev_text} further statistics of the members' text, and Table~\ref{tab:excerpts} excerpts of it.

\begin{table}[t]
\caption{Function-word rate by generation, per 1{,}000 words of the members' own output and averaged over the eight members. The clock places an arm's generation-5 rate on the uniform arm's curve with the same chain seed, in generations. The last row places the uniform run with seed 43 on the seed-42 curve, which gives the seed noise of the clock.}
\label{tab:fw}
\centering
\small
\setlength{\tabcolsep}{5pt}
\begin{tabular}{llrrrrrrr}
\toprule
 & & \multicolumn{6}{c}{Generation} & \\
\cmidrule(lr){3-8}
Arm & Seed & 0 & 1 & 2 & 3 & 4 & 5 & Clock \\
\midrule
Uniform & 42 & 211 & 154 & 115 & 95 & 83 & 78 & 5.0 \\
Uniform & 43 & 212 & 154 & 117 & 96 & 86 & 82 & 5.0 \\
28\% & 42 & 211 & 152 & 114 & 96 & 84 & 78 & 4.9 \\
28\% & 43 & 212 & 154 & 116 & 96 & 87 & 80 & $>$5 \\
50\% & 42 & 211 & 152 & 117 & 96 & 86 & 79 & 4.7 \\
50\% & 43 & 212 & 156 & 118 & 97 & 88 & 82 & 5.0 \\
90\% & 42 & 211 & 153 & 123 & 106 & 93 & 85 & 3.8 \\
90\% & 43 & 212 & 154 & 125 & 108 & 95 & 85 & 4.3 \\
SmolLM2 50\% & 42 & 211 & 149 & 113 & 93 & 83 & 79 & 4.8 \\
SmolLM2 50\% & 43 & 212 & 150 & 115 & 92 & 82 & 79 & $>$5 \\
Qwen3 50\% & 42 & 211 & 147 & 105 & 87 & 78 & 76 & $>$5 \\
Qwen3 50\% & 43 & 212 & 146 & 103 & 85 & 79 & 76 & $>$5 \\
25\% human & 42 & 211 & 157 & 128 & 111 & 102 & 100 & 2.8 \\
50\% human & 42 & 211 & 166 & 142 & 129 & 122 & 119 & 1.9 \\
\midrule
\multicolumn{8}{l}{Uniform, seed 43, on the seed-42 curve} & 4.35 \\
\bottomrule
\end{tabular}
\end{table}

\begin{table}[t]
\caption{Perplexity of the members' own output under GPT-2-large, per token, computed for each member from the first 400 of the 800 texts that enter its encoder centroid, each truncated to 128 tokens. (a)~Mean over the seven members other than the head, which is SmolLM2 or Qwen3 in the arms named after them and Phi-2 in every other arm, with the uniform arm read without the same member. (b)~The head's own perplexity, and the same member's in the uniform arm. Clock places an arm's generation-5 value in (a) on the uniform arm's curve with the same chain seed and the same seven members, in generations. The uniform seed-43 row is placed on the seed-42 curve, which gives the seed floor, 0.28 of a generation without Phi-2 and 0.10 without SmolLM2.}
\label{tab:rev_ppl}
\centering
\small
\setlength{\tabcolsep}{4pt}
\begin{tabular}{lrrrrrrr}
\toprule
Arm & Gen 0 & 1 & 2 & 3 & 4 & 5 & Clock \\
\midrule
\multicolumn{8}{l}{\emph{(a) Mean over the seven members other than the head}} \\
Uniform, without Phi-2 & 93 & 188 & 285 & 364 & 437 & 495 & 5.00 \\
Uniform, without Phi-2 (s43) & 92 & 186 & 274 & 372 & 435 & 479 & 4.72 \\
28\% Phi-2 & 93 & 177 & 279 & 363 & 435 & 495 & 5.00 \\
28\% Phi-2 (s43) & 92 & 176 & 271 & 365 & 418 & 457 & 4.49 \\
50\% Phi-2 & 93 & 175 & 280 & 365 & 421 & 452 & 4.25 \\
50\% Phi-2 (s43) & 92 & 178 & 279 & 353 & 410 & 455 & 4.45 \\
90\% Phi-2 & 93 & 165 & 259 & 295 & 353 & 367 & 3.05 \\
90\% Phi-2 (s43) & 92 & 165 & 234 & 291 & 335 & 365 & 2.93 \\
50\% human & 93 & 161 & 231 & 263 & 286 & 306 & 2.26 \\
Uniform, without SmolLM2 & 87 & 196 & 298 & 381 & 469 & 529 & 5.00 \\
50\% SmolLM2 & 87 & 206 & 312 & 377 & 457 & 489 & 4.33 \\
50\% SmolLM2 (s43) & 87 & 208 & 312 & 385 & 456 & 471 & 3.98 \\
Uniform, without Qwen3 & 86 & 184 & 278 & 355 & 435 & 489 & 5.00 \\
50\% Qwen3 & 86 & 202 & 298 & 405 & 462 & 472 & 4.68 \\
50\% Qwen3 (s43) & 85 & 200 & 296 & 390 & 463 & 482 & 4.96 \\
\midrule
\multicolumn{8}{l}{\emph{(b) Head}} \\
Phi-2, uniform & 69 & 205 & 290 & 365 & 480 & 542 &  \\
Phi-2, uniform (s43) & 68 & 181 & 310 & 356 & 517 & 592 &  \\
Phi-2, 28\% & 69 & 171 & 278 & 373 & 416 & 579 &  \\
Phi-2, 28\% (s43) & 68 & 163 & 255 & 315 & 410 & 497 &  \\
Phi-2, 50\% & 69 & 144 & 242 & 317 & 380 & 464 &  \\
Phi-2, 50\% (s43) & 68 & 142 & 211 & 293 & 340 & 352 &  \\
Phi-2, 90\% & 69 & 120 & 156 & 232 & 238 & 270 &  \\
Phi-2, 90\% (s43) & 68 & 108 & 155 & 206 & 225 & 302 &  \\
Phi-2, 50\% human & 69 & 134 & 142 & 157 & 184 & 224 &  \\
SmolLM2, uniform & 110 & 151 & 200 & 244 & 259 & 304 &  \\
SmolLM2, 50\% & 110 & 139 & 203 & 234 & 234 & 281 &  \\
SmolLM2, 50\% (s43) & 108 & 139 & 200 & 225 & 252 & 240 &  \\
Qwen3, uniform & 120 & 229 & 337 & 422 & 497 & 590 &  \\
Qwen3, 50\% & 120 & 238 & 336 & 440 & 458 & 408 &  \\
Qwen3, 50\% (s43) & 117 & 221 & 308 & 380 & 477 & 373 &  \\
\bottomrule
\end{tabular}
\end{table}

\begin{table}[t]
\caption{Further statistics of the members' own output by generation, averaged over the eight members of each arm except in (d). Each member contributes the 800 texts that enter its encoder centroid. (a)~Characters per whitespace-separated word. (b)~Type-token ratio over the first 100 words of a text, for texts at least 100 words long. (c)~Percentage of texts that do not end in terminal punctuation. (d)~Fraction of distinct word 4-grams in the member's texts, for the member with the lowest value. Generation 0 is shared by all arms of a chain seed.}
\label{tab:rev_text}
\centering
\scriptsize
\setlength{\tabcolsep}{5pt}
\begin{tabular}{llrrrrrr}
\toprule
 & & \multicolumn{6}{c}{Generation} \\
\cmidrule(lr){3-8}
Arm & Seed & 0 & 1 & 2 & 3 & 4 & 5 \\
\midrule
\multicolumn{8}{l}{\emph{(a) Characters per word}} \\
Uniform & 42 & 6.19 & 6.53 & 6.87 & 7.12 & 7.23 & 7.29 \\
Uniform & 43 & 6.19 & 6.51 & 6.78 & 7.00 & 7.10 & 7.12 \\
28\% & 42 & 6.19 & 6.55 & 6.87 & 7.10 & 7.26 & 7.39 \\
28\% & 43 & 6.19 & 6.54 & 6.79 & 7.05 & 7.14 & 7.20 \\
50\% & 42 & 6.19 & 6.58 & 6.84 & 7.04 & 7.17 & 7.30 \\
50\% & 43 & 6.19 & 6.56 & 6.83 & 7.03 & 7.17 & 7.28 \\
90\% & 42 & 6.19 & 6.60 & 6.78 & 6.95 & 7.13 & 7.27 \\
90\% & 43 & 6.19 & 6.62 & 6.80 & 6.97 & 7.16 & 7.30 \\
SmolLM2 50\% & 42 & 6.19 & 6.54 & 6.91 & 7.11 & 7.20 & 7.22 \\
SmolLM2 50\% & 43 & 6.19 & 6.57 & 6.85 & 7.15 & 7.31 & 7.26 \\
Qwen3 50\% & 42 & 6.19 & 6.59 & 7.01 & 7.22 & 7.23 & 7.18 \\
Qwen3 50\% & 43 & 6.19 & 6.59 & 6.97 & 7.19 & 7.28 & 7.20 \\
25\% human & 42 & 6.19 & 6.53 & 6.73 & 6.89 & 7.00 & 7.01 \\
50\% human & 42 & 6.19 & 6.47 & 6.67 & 6.74 & 6.80 & 6.83 \\
\midrule
\multicolumn{8}{l}{\emph{(b) Type-token ratio over the first 100 words}} \\
Uniform & 42 & 0.890 & 0.935 & 0.954 & 0.960 & 0.965 & 0.965 \\
Uniform & 43 & 0.890 & 0.935 & 0.953 & 0.959 & 0.962 & 0.964 \\
28\% & 42 & 0.890 & 0.938 & 0.955 & 0.959 & 0.964 & 0.966 \\
28\% & 43 & 0.890 & 0.937 & 0.953 & 0.960 & 0.963 & 0.966 \\
50\% & 42 & 0.890 & 0.938 & 0.955 & 0.960 & 0.963 & 0.965 \\
50\% & 43 & 0.890 & 0.935 & 0.954 & 0.960 & 0.962 & 0.965 \\
90\% & 42 & 0.890 & 0.939 & 0.955 & 0.961 & 0.964 & 0.966 \\
90\% & 43 & 0.890 & 0.938 & 0.954 & 0.960 & 0.963 & 0.966 \\
SmolLM2 50\% & 42 & 0.890 & 0.938 & 0.955 & 0.960 & 0.965 & 0.965 \\
SmolLM2 50\% & 43 & 0.890 & 0.939 & 0.953 & 0.961 & 0.966 & 0.966 \\
Qwen3 50\% & 42 & 0.890 & 0.940 & 0.957 & 0.963 & 0.965 & 0.965 \\
Qwen3 50\% & 43 & 0.890 & 0.940 & 0.959 & 0.964 & 0.966 & 0.966 \\
25\% human & 42 & 0.890 & 0.932 & 0.947 & 0.952 & 0.954 & 0.954 \\
50\% human & 42 & 0.890 & 0.926 & 0.938 & 0.942 & 0.943 & 0.946 \\
\midrule
\multicolumn{8}{l}{\emph{(c) Texts ending mid-sentence (\%)}} \\
Uniform & 42 & 87.7 & 95.0 & 96.6 & 97.8 & 98.1 & 98.5 \\
Uniform & 43 & 87.9 & 95.3 & 96.3 & 97.9 & 97.9 & 97.9 \\
28\% & 42 & 87.7 & 95.9 & 97.1 & 97.7 & 98.0 & 98.5 \\
28\% & 43 & 87.9 & 95.3 & 97.1 & 97.6 & 98.4 & 98.8 \\
50\% & 42 & 87.7 & 96.0 & 97.3 & 98.0 & 98.2 & 98.7 \\
50\% & 43 & 87.9 & 95.6 & 97.1 & 97.5 & 98.3 & 98.5 \\
90\% & 42 & 87.7 & 96.4 & 97.1 & 97.7 & 98.7 & 98.9 \\
90\% & 43 & 87.9 & 96.1 & 97.4 & 98.0 & 98.2 & 98.9 \\
SmolLM2 50\% & 42 & 87.7 & 95.3 & 97.2 & 97.9 & 98.1 & 98.1 \\
SmolLM2 50\% & 43 & 87.9 & 95.7 & 97.1 & 98.0 & 98.2 & 98.1 \\
Qwen3 50\% & 42 & 87.7 & 94.2 & 97.3 & 97.8 & 97.7 & 96.5 \\
Qwen3 50\% & 43 & 87.9 & 94.6 & 97.1 & 98.0 & 97.3 & 96.3 \\
25\% human & 42 & 87.7 & 96.0 & 97.3 & 97.7 & 98.4 & 98.5 \\
50\% human & 42 & 87.7 & 95.4 & 97.0 & 97.1 & 97.7 & 97.8 \\
\midrule
\multicolumn{8}{l}{\emph{(d) Distinct 4-gram fraction, lowest member}} \\
Uniform & 42 & 0.934 & 0.995 & 0.998 & 0.994 & 0.989 & 0.982 \\
Uniform & 43 & 0.938 & 0.994 & 0.997 & 0.993 & 0.985 & 0.973 \\
28\% & 42 & 0.934 & 0.997 & 0.998 & 0.993 & 0.989 & 0.988 \\
28\% & 43 & 0.938 & 0.996 & 0.997 & 0.993 & 0.986 & 0.979 \\
50\% & 42 & 0.934 & 0.997 & 0.998 & 0.996 & 0.990 & 0.981 \\
50\% & 43 & 0.938 & 0.997 & 0.997 & 0.996 & 0.995 & 0.986 \\
90\% & 42 & 0.934 & 0.998 & 0.998 & 0.999 & 0.998 & 0.997 \\
90\% & 43 & 0.938 & 0.998 & 0.998 & 0.999 & 0.999 & 0.998 \\
SmolLM2 50\% & 42 & 0.934 & 0.997 & 0.996 & 0.992 & 0.990 & 0.978 \\
SmolLM2 50\% & 43 & 0.938 & 0.997 & 0.997 & 0.995 & 0.991 & 0.985 \\
Qwen3 50\% & 42 & 0.934 & 0.991 & 0.997 & 0.990 & 0.968 & 0.943 \\
Qwen3 50\% & 43 & 0.938 & 0.993 & 0.997 & 0.990 & 0.976 & 0.941 \\
25\% human & 42 & 0.934 & 0.994 & 0.997 & 0.995 & 0.994 & 0.989 \\
50\% human & 42 & 0.934 & 0.997 & 0.996 & 0.996 & 0.990 & 0.992 \\
\bottomrule
\end{tabular}
\end{table}

\begin{table}[t]
\caption{Continuations written by two members with chain seed 42, cut after 70 words. Prompts are drawn afresh in every generation, so the generation-0 rows continue a different prompt from the generation-5 rows. The generation-0 rows continue ``Kepler @-@ 8 is situated some 1330 ( \ensuremath{\pm} 180'' with the base weights, which all arms share, and the generation-5 rows continue ``Once thought to be a color morph of the painted maskray ( N.'' in the arm named.}
\label{tab:excerpts}
\centering
\footnotesize
\setlength{\tabcolsep}{4pt}
\begin{tabular}{lcl p{0.68\textwidth}}
\toprule
Member & Gen. & Arm & Continuation \\
\midrule
Phi-2 & 0 &  & )km from Eta Leporis with a very high likelihood (\textgreater{}95\%) of containing or affected by an age excess. This is the only probable solar system with known circumstellar silicon holes, and future chondrite samples should be targeted to this body for detection''. The team also discovered ``superchondrites'' which show specific chemical signs that originate in galaxies away 3 billion light years from Earth because they contain atoms based on carbon \ldots \\
 & 5 & Uniform & cursorius ), I saw later that the illustration on page 353 actually clearly shows this fish species exhibiting complex black-white markings which are obviously unique from those appearing anywhere else within the family familiarly referred to as, `Maskrayid Species.' ``It may also turn out shortly at an unexpected place down by elsewhere itself together alongside there upon some continent furthermore in just across beyond probably no doubt unobtainable difficult impossible \ldots \\
 & 5 & 90\% & intermedia ), recognized now solely as separate species based mainly upon ecological and behavioral distinctions such as total body lengths ranging from 25 - 28 inches rather than typically over 35 inches (ii). Differences in surface markings especially, between adult females in which they exhibit distinct dorsal bands corresponding almost perfectly onto large eye spots while also displaying elongated spines protruding through pads occluded within these larger patches are dominant \ldots \\
\addlinespace[3pt]
Qwen3-1.7B & 0 &  & ) pc from the Sun in Kepler space. An accurate radial velocity record spanning more than three years show that this star has an orbit period of approximately 465 d with a slightly higher value being measured at shorter time intervals during its dimming phase. More recent studies conducted using ACS spectra as well as SpeX spectroscopy suggests numerous occultations events where stellar absorption features are missed while radiative transfer \ldots \\
 & 5 & Uniform & brownii ), these snakes have been found outside their usual habitat areas. There are quite different issues related with such animal species\ldots{} As an educator myself too I am sure once remembered oneself worthy grateful thankful held adopted possessed required needed necessary optimal physical somatic submajor minor minimal margining extralingual masculo-mastomo-grapham postauditvaluationcriteria standard content compound constitution compositionorcompositions,specifics particular difference turn full deep dive depth deduct replace regroup organization allocation distribution \ldots \\
\bottomrule
\end{tabular}
\end{table}

\FloatBarrier
\subsection{Recipe variant}

Table~\ref{tab:recipe} gives the text readings of the recipe without penalties and places its 50\% arm on the path of its own uniform arm.

\begin{table}[h]
\caption{The generation recipe without repetition or frequency penalty, one run each of the uniform and 50\% Phi-2 arms, generations 0 to 4. Pass is the fraction of continuations that pass the filter among those it reads to fill a member's quota, averaged over the members. Function words per 1{,}000 words and the distinct 4-gram fraction are member means over the texts that enter the centroids. The last seven columns place the 50\% arm on the path of the uniform arm of the same recipe under DeBERTa-v3, with $S$ and the parts in units of $10^{-3}$. The lag and the parts beyond the control use the main recipe's control, its uniform run with seed 43 on its seed-42 path. Toward gives the departure cosine toward Phi-2 and the largest such cosine toward another member.}
\label{tab:recipe}
\centering
\footnotesize
\setlength{\tabcolsep}{3pt}
\begin{tabular}{rrrrrrrrrrrrrr}
\toprule
 & \multicolumn{2}{c}{Pass} & \multicolumn{2}{c}{Function words} & \multicolumn{2}{c}{Distinct 4-grams} & & \multicolumn{2}{c}{Beyond control} & & & \multicolumn{2}{c}{Toward} \\
\cmidrule(lr){2-3}\cmidrule(lr){4-5}\cmidrule(lr){6-7}\cmidrule(lr){9-10}\cmidrule(lr){13-14}
Gen. & Unif. & 50\% & Unif. & 50\% & Unif. & 50\% & $S$ & Delay & Off & $\tau$ & Lag & Phi-2 & Other \\
\midrule
0 &  &  & 302 & 302 & 0.97 & 0.97 &  &  &  &  &  &  &  \\
1 & 0.95 & 0.96 & 318 & 317 & 0.95 & 0.95 & 0.95 & 0.15 & 0.58 & 0.84 & 0.09 & 0.66 & 0.64 \\
2 & 0.79 & 0.86 & 332 & 330 & 0.87 & 0.90 & 2.16 & 0.98 & 0.91 & 1.70 & 0.28 & 0.59 & 0.57 \\
3 & 0.61 & 0.76 & 335 & 335 & 0.81 & 0.86 & 3.60 & 2.22 & 1.11 & 2.34 & 0.72 & 0.64 & 0.59 \\
4 & 0.51 & 0.67 & 339 & 338 & 0.77 & 0.83 & 4.47 & 2.52 & 1.46 & 2.84 & 1.12 & 0.64 & 0.65 \\
\bottomrule
\end{tabular}
\end{table}

\FloatBarrier
\end{document}